\documentclass{article}

     \usepackage[preprint]{neurips_2019}

\usepackage[utf8]{inputenc} 
\usepackage[T1]{fontenc}    
\usepackage{hyperref}       
\usepackage{url}            
\usepackage{booktabs}       
\usepackage{amsfonts}       
\usepackage{nicefrac}       
\usepackage{microtype}      

\usepackage[ruled,linesnumbered]{algorithm2e}
\usepackage{amssymb}
\usepackage{amsmath}
\usepackage{amsthm}
\usepackage{bbm}
\usepackage{enumitem}
\usepackage{mathtools}
\usepackage{xcolor}

\theoremstyle{definition}

\newtheorem{lemma}{Lemma}

\newtheorem{proposition}{Proposition}

\renewcommand\thelemma{\arabic{lemma}}
\renewcommand\theproposition{\arabic{proposition}}

\newcommand{\dd}{\text{d}}
\newcommand{\RR}{\mathbb{R}}
\newcommand{\pl}{\text{pl}}

\newcommand{\fpartial}[2]{\frac{\partial #1}{\partial #2}}

\newcommand{\Var}{\text{var}}

\newcommand{\Cov}{\text{cov}}
\newcommand{\E}{\text{E}}
\newcommand{\pQ}{\text{pQ}}

\newcommand{\sumjm}{\sum_{j=1}^3}
\newcommand{\sumjminj}{\sum_{j=1}^3\sum_{i=1}^{n_j}}

\def\T{ {\mathrm{\scriptscriptstyle T}} }
\def\me{\mathrm e}

\title{Semi-supervised Logistic Learning Based on Exponential Tilt Mixture Models}

\author{%
  Xinwei Zhang \, \& \, Zhiqiang Tan \\
  Department of Statistics, Rutgers University, USA \\
  \texttt{xinwei.zhang@rutgers.edu, ztan@stat.rutgers.edu} \\
}

\begin{document}

\maketitle

\begin{abstract}
Consider semi-supervised learning for classification, where both labeled and unlabeled data are available for training.
The goal is to exploit both datasets to achieve higher prediction accuracy than just using labeled data alone.
We develop a semi-supervised logistic learning method based on exponential tilt mixture models, by
extending a statistical equivalence between logistic regression and exponential tilt modeling.
We study maximum nonparametric likelihood estimation and derive novel objective functions which are shown to be Fisher consistent.
We also propose regularized estimation and construct simple and highly interpretable EM algorithms.
Finally, we present numerical results which demonstrate the advantage of the proposed methods compared with existing methods.
\end{abstract}

\section{Introduction}

Semi-supervised learning for classification involves exploiting a large amount of unlabeled data and a relatively small amount of labeled data
to build better classifiers. This approach can potentially be used to achieve higher accuracy, with a limited budget for obtaining labeled data.
Various methods have been proposed, including expectation-maximization (EM) algorithms, transductive support vector machines (SVMs),
and regularized methods (e.g., Chapelle et al.~2006; Zhu 2008).

For supervised classification, there are a range of objective functions
which are Fisher consistent in the following sense: optimization of the population, nonparametric version of a loss function leads to
the true conditional probability function of labels given features as for the logistic loss,
or to the Bayes classifier as for the hinge loss (Lin 2002; Bartlett et al.~2006). 
In contrast, a perplexing issue we notice for semi-supervised classification is that existing objective functions are in general not Fisher consistent,
unless in the degenerate case where unlabeled data are ignored and only labeled data are used.
Examples include the objective functions in transductive SVMs (Vapnik 1998; Joachims 1999) and
various regularized methods (Grandvalet \& Bengio 2005; Mann \& McCallum 2007; Krishnapuram et al.~2005).
The lack of Fisher consistency may contribute to unstable performances of existing semi-supervised classifiers (e.g., Li \& Zhou 2015).
Another restriction in existing methods is that the class proportions in labeled and unlabeled data are typically assumed to be the same.

We develop a semi-supervised extension of logistic regression based on exponential tilt mixture models (Qin 1999; Zou et al.~2002; Tan 2009),
{\it without} restricting the class proportions in the unlabeled data to be the same as in the labeled data.
The development is motivated by a statistical equivalence between logistic regression for the conditional probability of a label given features and
exponential tilt modeling for the density ratio between the feature distributions within different labels (Anderson 1972; Prentice \& Pyke 1979).
Our work involves two main contributions: (i) we derive novel objective functions which are shown not only to be
Fisher consistent but also lead to asymptotically more efficient estimation than based on labeled data only,
and (ii) we propose regularized estimation and construct computationally and conceptually desirable EM algorithms.
From numerical experiments, our methods achieve a substantial advantage over existing methods when the class proportions in unlabeled data differ from those in labeled data.
A possible explanation is that while the class proportions in unlabeled data are estimated as unknown parameters in our methods, they are
implicitly assumed to be the same as in labeled data for existing methods including transductive SVMs (Joachims 1999) and entropy regularization (Grandvalet \& Bengio 2005).

A simple, informative example is provided in the Supplement (Section~\ref{sec:illustration}) to highlight comparison between new and existing methods mentioned above.

\section{Background: logistic regression and exponential tilt model} \label{sec:logistic}

For supervised classification, the training data consist of a sample $\{(y_i, x_i): i=1,\ldots, n\}$ of $(y,x)$, where $x \in \RR^p$ and $y \in \{0,1\}$
representing a feature vector and an associated label respectively.
Consider a logistic regression model \vspace{-.1in}
\begin{align}
    \label{fm:logistic_regression}
      P(y =1 |x ) = \frac{\exp(\beta^c_0 +\beta_1^\T x)}{1+\exp(\beta^c_0 +\beta_1^\T x)} ,
\end{align}
where $\beta_1$ is a coefficient vector associated with $x$, and
$\beta_0^c$ is an intercept, with superscript $^c$ indicating classification or conditional probability of $y=1$ given $x$.
The maximum likelihood estimator (MLE) $(\tilde \beta_0^c, \tilde \beta_1)$ is defined as a maximizer of the log (conditional) likelihood: \vspace{-.1in}
\begin{align}
\sum_{i=1}^n \Big[ y_i (\beta^c_0 +\beta_1^\T x_i) - \log \{ 1+\exp(\beta^c_0 +\beta_1^\T x_i) \} \Big] . \label{eq:logistic-lik}
\end{align}
In general, nonlinear functions of $x$ can be used in place of $\beta_1^\T x$, and a penalty term can be incorporated into the log-likelihood such as the ridge penalty $\|\beta_1\|_2^2$ or
the squared norm of a reproducing kernel Hilbert space of functions of $x$. We discuss these issues later in Sections ~\ref{sec:EM} and \ref{sec:conclusion}.

Interestingly, logistic regression on $P(y|x)$ can be made equivalent to an exponential tilt model on $P(x|y)$ (Anderson 1972; Prentice \& Pyke 1979; Qin 1998).
Denote by $G_0$ or $G_1$ the conditional distribution $P(x|y=0)$ or $P(x|y=1)$ respectively, and $\pi=P(y=1)$.
By the Bayes rule, model (\ref{fm:logistic_regression}) is equivalent to the exponential tilt model
\begin{align}
\frac{\dd P(x|y=1)}{\dd P(x|y=0)}= \frac{\dd G_1}{\dd G_0} = \me^{\beta_0 + \beta_1^\T x}, \label{fm:exp-tilt}
\end{align}
where $\dd G_1/\dd G_0$ denotes the density ratio between $G_1$ and $G_0$ with respect to a dominating measure, and $\beta_0 = \beta_0^c + \log\{(1-\pi)/\pi\}$.
Model (\ref{fm:exp-tilt}) is explicitly a semi-parametric model, where $G_0$ is an infinitely-dimensional parameter
and $(\beta_0, \beta_1)$ are finitely-dimensional parameters. In fact, logistic model (\ref{fm:logistic_regression}) is also semi-parametric, where
the marginal distribution of $x$ is an infinitely-dimensional parameter, and $(\beta_0^c,\beta_1)$ are finitely-dimensional parameters.
Furthermore, the MLE $(\tilde \beta_0^c, \tilde \beta_1)$ in model (\ref{fm:logistic_regression}) can be related to the following estimator $(\hat\beta_0, \hat\beta_1)$ in model (\ref{fm:exp-tilt})
by the method of nonparametric likelihood (Kiefer \& Wolfowitz 1956) or empirical likelihood (Owen 2001).
Formally,  $(\hat\beta_0, \hat\beta_1, \hat G_0)$ are defined as a maximizer of the log-likelihood, \vspace{-.1in}
\begin{align}
\sum_{i=1}^n \Big[ (1-y_i) + y_i (\beta_0 + \beta_1^\T x_i) + \log G_0(\{x_i\}) \Big], \label{eq:ET-lik}
\end{align}
over all possible $(\beta_0,\beta_1,G_0)$ such that $G_0$ is a probability measure supported on the pooled data $\{x_i : i=1,\ldots,n\}$ with
$\int \exp(\beta_0 + \beta_1^\T x) \dd G_0 =1$.
Analytically, it can be shown that
$ \tilde\beta_1 = \hat\beta_1$, $\tilde\beta_0^c = \hat \beta_0 + \log\{\hat\pi/(1-\hat\pi)\}$,
where $\hat \pi = \sum_{i=1}^n y_i /n$. See Qin (1998) and references therein.

By the foregoing discussion, we see that there are two statistically distinct but equivalent approaches for supervised classification: logistic regression
or exponential tilt models. It is such a relationship that we aim to exploit in developing a new method for semi-supervised classification.


\section{Theory and methods}

For semi-supervised classification, the training data consist of a labeled sample $\mathcal S^\ell=\{(y_i, x_i): i=1,\ldots, n\}$ and
an unlabeled sample $\mathcal S^u=\{x_i : i = n+1, \ldots, N\}$, for which the associated labels $\{y_i: i=n+1,\ldots, N\}$ are unobserved.
Typically for existing methods including transductive SVMs, the two samples $\mathcal S^\ell$ and $\mathcal S^u$  are assumed to be from a common population of $(y,x)$.
However, we allow that $\mathcal S^\ell$ and $\mathcal S^u$ may be drawn from different populations, with the same
conditional distribution $P(x|y)$, but possibly different marginal probabilities $P^\ell(y=1)$ and $P^u(y=1)$.

\subsection{Exponential tilt mixture model} \label{sec:ETM}

Although it seems difficult at first look to extend logistic model (\ref{fm:logistic_regression}) for semi-supervised learning,
we realize that both the labeled sample $\mathcal S^\ell$ and
the unlabeled sample $\mathcal S^u$ can be taken account of by a  natural extension of the exponential tilt model (\ref{fm:exp-tilt}),
called an exponential tilt mixture (ETM) model (Qin 1999; Zou et al.~2002; Tan 2009). Denote
\begin{align*}
& \mathcal S_1 = \{x_i : y_i =0, i=1,\ldots,n\} \mbox{ drawn from } P_1 (x) = P(x|y=0),\\
& \mathcal S_2 = \{x_i : y_i =1, i=1,\ldots,n\} \mbox{ drawn from } P_2 (x) = P(x|y=1), \\
& \mathcal S_3 =  \{x_i : i = n+1, \ldots, N \} \quad \mbox{ drawn from } P_3 (x) = P^u(x) .
\end{align*}
An exponential tilt mixture model for the three samples $(\mathcal S_1,\mathcal S_2,\mathcal S_3)$ postulates that
\begin{align}
& \dd P_1(x) = \dd G_0(x), \label{eq:ETM1}\\
& \dd P_2(x) = \dd G_1(x), \label{eq:ETM2} \\
& \dd P_3(x) = (1-\rho) \dd G_0(x) + \rho \,\dd G_1(x), \label{eq:ETM3}
\end{align}
where $G_0$ or $G_1$ represents the conditional distribution of $x$ given $y=0$ or $y=1$ respectively in both the labeled and unlabeled data such that
\begin{align}
\frac{\dd G_1}{\dd G_0} = \me^{\beta_0 + \beta_1^\T x}, \label{fm:exp-tilt2}
\end{align}
and $\rho = P^u(y=1)$ is the proportion of $y=1$ underlying the unlabeled data.
While Eqs~(\ref{eq:ETM1})--(\ref{eq:ETM2}) merely give definitions of $G_0$ and $G_1$,
Eq~(\ref{eq:ETM3}) says that the feature distribution in the unlabeled sample is a mixture of $G_0$ and $G_1$, which follows from the structural assumption that
the conditional distribution $P(x|y)$ is invariant between the labeled and unlabeled samples.
Eq~(\ref{fm:exp-tilt2}) imposes a functional restriction on the density ratio between $G_0$ and $G_1$, similarly as in (\ref{fm:exp-tilt}).

The ETM model, defined by (\ref{eq:ETM1})--(\ref{fm:exp-tilt2}), is a semi-parametric model, with an infinitely-dimensional parameter $G_0$
and finitely-dimensional parameter $\rho$ and $\beta=(\beta_0,\beta_1^\T)^\T$. We briefly summarize maximum nonparametric likelihood estimation
previously studied (Qin 1999; Zou et al.~2002; Tan 2009).
For notational convenience, rewrite the sample $\mathcal S_j$ as $\{x_{ji}: i=1,\ldots, n_j\}$, where $n_1 = n-n_2$, $n_2 = \sum_{i=1}^n y_i$, and $n_3 = N-n$.
Eqs~(\ref{eq:ETM1})--(\ref{eq:ETM3}) can be expressed as
\begin{align*}
   \dd P_j = (1-\rho_j) \dd G_0 + \rho_j \,\dd G_1, \quad j=1,2,3,
\end{align*}
where $\rho_1= 0$, $\rho_2=1$, and $\rho_3 = \rho$.
For any fixed $(\rho,\beta)$, the average profile log-likelihood of $(\rho,\beta)$ is defined as
$\pl(\rho,\beta) = \max_{G_0} l(\rho,\beta, G_0)$ with
\begin{align}
    l (\rho, \beta,  G_0) = \frac{1}{N}   \sumjminj \Big[ \log \{1-\rho_j+ \rho_j \exp(\beta_0+\beta_1^\T x_{ji})\} + \log  G_0(\{x_{ji}\}) \Big], \label{eq:full-lik}
\end{align}
over all possible $G_0$ which is a probability measure supported on the pooled data $\{x_{ji} : i=1,\ldots,n_j, j=1,2,3\}$ with
$\int \exp(\beta_0 + \beta_1^\T x) \dd G_0 =1$.
Denote
\begin{align*}
 \kappa( \rho, \beta, \alpha) =  \frac{1}{N}  \sumjminj \log \left\{ \frac{ 1-\rho_j + \rho_j \exp(\beta_0+ \beta_1^\T x_{ji})}{ 1- \alpha + \alpha \exp(\beta_0+\beta_1^\T x_{ji})} \right\} - \log N,
\end{align*}
which can be easily shown to be concave in $\rho \in (0,1)$ and convex in $\alpha \in (0,1)$.
Then Proposition~1 in Tan (2009) leads to the following result.

\begin{lemma} \label{lem:profile}
The average profile log-likelihood of $(\rho,\beta)$  can be determined as
$\pl(\rho,\beta) = \min_{\alpha \in(0,1)} \kappa ( \rho, \beta, \alpha) = \kappa \{ \rho, \beta, \hat\alpha(\beta)\}$,
where $\hat\alpha(\beta)$ is a minimizer of $\kappa ( \rho, \beta, \alpha)$ over $\alpha$, satisfying the stationary condition (free of $\rho$)
\begin{align}
    1 = \frac{1}{N}\sumjminj \frac{1}{1-\alpha+\alpha\exp(\beta_0+\beta_1^\T x_{ji})}. \label{eq:alpha}
\end{align}
\end{lemma}

The maximum likelihood estimator of $(\rho,\beta)$ is then defined by maximizing the profile log-likelihood, that is, $(\hat\rho,\hat\beta) = \mbox{argmax}_{(\rho,\beta)} \pl(\rho,\beta)$.
From Lemma~\ref{lem:profile}, we notice that the estimators $\{\hat\rho, \hat\beta, \hat\alpha(\hat\beta)\}$ jointly solve the saddle-point problem:
\begin{align}
\max_{(\rho,\beta)} \min_\alpha \, \kappa ( \rho, \beta, \alpha) . \label{eq:saddle-point}
\end{align}
Large sample theory of $(\hat\rho,\hat\beta)$ has been studied in Qin (1999) under standard regularity conditions as $N\to\infty$ and $n_j/N\to\eta_j$ with some constant $\eta_j >0$ for $j=1,2,3$.
The theory shows the existence of a local maximizer of $\pl(\rho,\beta)$, which is consistent and asymptotically normal provided the ETM model (\ref{eq:ETM1})--(\ref{fm:exp-tilt2})
is correctly specified. However, there remain subtle questions. It seems unclear whether the population version of the average profile log-likelihood $\kappa \{ \rho, \beta, \hat\alpha(\beta)\}$
attains a global maximum at the true values of $(\rho,\beta)$ under a correctly specified ETM model.
Moreover, what property can be deduced for $(\hat\rho,\hat\beta)$ under a misspecified ETM model?

\subsection{Semi-supervised logistic regression} \label{sec:SLR}

We derive a new classification model with parameters $(\rho,\beta)$ for the three samples  $(\mathcal S_1,\mathcal S_2,\mathcal S_3)$  such that
an MLE of $(\rho,\beta)$ in the new model coincides with an MLE $(\hat \rho,\hat\beta)$ in the ETM model, and vice versa.
Let $z_i=1+y_i$ if $i=1, \ldots, n$ and $z_i = 3$ if $i = n+1, \ldots, N$. Consider a conditional probability model for predicting the label $z_i$ from $x_i$:
\begin{align}
& P( z=1 | x) = \frac{n_1}{N} \frac{1}{ 1-\tilde\alpha(\rho) + \tilde\alpha(\rho) \exp(\beta_0+\beta_1^\T x)} , \label{eq:new-logit1}\\
& P( z=2 | x) = \frac{n_2}{N} \frac{\exp(\beta_0+\beta_1^\T x)}{ 1-\tilde\alpha(\rho) + \tilde\alpha(\rho) \exp(\beta_0+\beta_1^\T x)} , \label{eq:new-logit2} \\
& P( z=3 | x) = \frac{n_3}{N} \frac{1-\rho + \rho \exp(\beta_0+\beta_1^\T x)}{ 1-\tilde\alpha(\rho) + \tilde\alpha(\rho) \exp(\beta_0+\beta_1^\T x)} ,  \label{eq:new-logit3}
\end{align}
where $\tilde\alpha(\rho) = \sum_{j=1}^3 (n_j/N) \rho_j = (n_2 + n_3 \rho)/N$, which ensures that $\sum_{j=1}^3 P(z=j|x) \equiv 1$.
The model, defined by (\ref{eq:new-logit1})--(\ref{eq:new-logit3}), will be called a semi-supervised logistic regression (SLR) model.
The average log-likelihood function of $(\rho, \beta)$ with the data $\{(z_i,x_i): i=1,\ldots,N\}$ in model (\ref{eq:new-logit1})--(\ref{eq:new-logit3}) can be written,
up to an additive constant free of $(\rho,\beta)$, as
\begin{align*}
 \kappa\{ \rho, \beta, \tilde\alpha(\rho)\} =  \frac{1}{N}
 \sumjminj \log \left\{ \frac{ 1-\rho_j + \rho_j \exp(\beta_0+ \beta_1^\T x_{ji})}{ 1- \tilde\alpha(\rho) + \tilde\alpha(\rho) \exp(\beta_0+\beta_1^\T x_{ji})} \right\} - \log N .
\end{align*}

\begin{proposition} \label{prop:equiv}
If and only if $(\hat\rho,\hat\beta)$ is a local (or respectively global) maximizer of the average log-likelihood $\kappa\{ \rho, \beta, \tilde\alpha(\rho) \}$ in SLR model (\ref{eq:new-logit1})--(\ref{eq:new-logit3}),
then it is a local (or global) maximizer of the average profile log-likelihood $\kappa \{ \rho, \beta, \hat\alpha(\beta)\}$ in ETM model (\ref{eq:ETM1})--(\ref{fm:exp-tilt2}).
\end{proposition}

Proposition~\ref{prop:equiv} shows an equivalence between maximum {\it nonparametric} likelihood estimation
in ETM model (\ref{eq:ETM1})--(\ref{fm:exp-tilt2})
and {\it usual} maximum likelihood estimation in SLR model (\ref{eq:new-logit1})--(\ref{eq:new-logit3}), even though
the objective functions $\kappa \{ \rho, \beta, \hat\alpha(\beta)\}$ and $\kappa\{ \rho, \beta, \tilde\alpha(\rho)\}$ are not equivalent.
This differs from the equivalence between logistic regression (\ref{fm:logistic_regression}) and exponential tilt model (\ref{fm:exp-tilt}) with labeled data only,
where the log-likelihood (\ref{eq:logistic-lik}) and the profile log-likelihood from (\ref{eq:ET-lik}) are equivalent (Prentice \& Pyke 1979).
From another angle, this result says that saddle-point problem (\ref{eq:saddle-point}) can be equivalently solved by directly maximizing $\kappa\{\rho, \beta, \tilde\alpha(\rho)\}$.
This transformation is nontrivial, because a saddle-point problem in general cannot be converted into optimization with a closed-form objective.

By the identification of $\kappa\{ \rho, \beta, \tilde\alpha(\rho) \}$ as a usual log-likelihood function,
we show that the objective functions $\kappa \{ \rho, \beta, \hat\alpha(\beta)\}$ and $\kappa\{ \rho, \beta, \tilde\alpha(\rho)\}$, with the linear predictor $\beta_0 + \beta_1^\T x$
replaced by an arbitrary function $h(x)$, are Fisher consistent nonparametrically, i.e., maximization of their population versions leads to the true values.
This seems to be the first time Fisher consistency of a loss function is established for semi-supervised classification.
By some abuse of notation, denote
\begin{align*}
 \kappa( \rho, h, \alpha) =  \frac{1}{N}  \sumjminj \log \left\{ \frac{ 1-\rho_j + \rho_j \exp(h(x_{ji}))}{ 1- \alpha + \alpha \exp(h( x_{ji}))} \right\} - \log N.
\end{align*}

\begin{proposition} \label{prop:consistency}
Suppose that $\mathcal S_j = \{x_{ji}: i=1,\ldots, n_j\}$ is drawn from $P_j$ in (\ref{eq:ETM1})--(\ref{eq:ETM3}) for $j=1,2,3$, with $\rho=\rho^*$
and $\dd G_1 / \dd G_0 = \exp(h^*(x))$ for some fixed value $\rho^*\in (0,1)$ and function $h^*(x)$. Denote
$  \kappa^*( \rho, h, \alpha)  =  E \{ \kappa( \rho, h, \alpha)  \}$.
For any $\rho\in (0,1)$ and function $h(x)$, we have
\begin{align*}
\min_{\alpha\in(0,1)} \kappa^*(\rho, h, \alpha) \le \kappa^*\{\rho, h, \tilde \alpha(\rho) \} \le \kappa^*\{ \rho^*, h^*, \tilde \alpha(\rho^*) \} ,
\end{align*}
where both equalities hold if $\rho=\rho^*$ and $h=h^*$.
Hence the population objective functions $\kappa^*\{ \rho, h, \tilde \alpha(\rho)\}$ and $\min_{\alpha\in(0,1)} \kappa^*(\rho, h, \alpha)$ are maximized at the true value $\rho^*$ and function $h^*(x)$.
\end{proposition}

Proposition~\ref{prop:consistency} fills existing gaps in understanding maximum likelihood estimation in ETM model (\ref{eq:ETM1})--(\ref{fm:exp-tilt2}),
through its equivalence with that in SLR model (\ref{eq:new-logit1})--(\ref{eq:new-logit3}). If the ETM model is correctly specified, then
the population version of $\kappa \{ \rho, \beta, \hat\alpha(\beta)\}$ has a global maximum at the true values of $(\rho,\beta)$, and hence
a global maximizer $(\hat\rho,\hat\beta)$ is consistent under suitable regularity conditions.
If the ETM model is misspecified, then by theory of estimation with misspecified models (Manski 1988; White 1982), the MLE $(\hat\rho,\hat\beta)$ converges in probability to a limit value which minimizes the difference between
$\kappa^*\{\rho, h, \tilde \alpha(\rho) \}$ with $h=\beta_0+\beta_1^\T x$ and $\kappa^*\{ \rho^*, h^*, \tilde \alpha(\rho^*) \}$. This difference as shown in the Supplement (Section~\ref{sec:prf-KL}) is
the expected Kullback--Leibler divergence
\begin{align*}
E \left\{ \mbox{KL} \Big(w(\cdot,x; \rho^*,h^*) \| w(\cdot,x; \rho,h) \Big) \right\},
\end{align*}
where $w(j,x; \rho, h)$ is the conditional probability (\ref{eq:new-logit1})--(\ref{eq:new-logit3}) for $j=1,2,3$,
$\mbox{KL}( q^* \| q )$ is the Kullback--Leibler divergence between two probability vectors $(q^*_j)$ and $(q_j)$, and
$E(\cdot)$ denotes the expectation with respect to $(1-\tilde\alpha(\rho^*)) \dd G_0 +\tilde\alpha(\rho^*) \dd G_1$.


Finally, we point out another interesting property of SLR model (\ref{eq:new-logit1})--(\ref{eq:new-logit3}).
If $\rho$ is fixed as $\rho^\ell= n_2/n$, the proportion of $y=1$ in the labeled sample, then
$\tilde\alpha(\rho^\ell) =  (n_2 + n_3 \rho^\ell)/N = n_2/n$.
In this case, the conditional probability (\ref{eq:new-logit3}) reduces to a constant, and
the objective function $\kappa\{ \rho^\ell, \beta, \tilde\alpha(\rho^\ell)\}$ can be easily shown to be equivalent to
the profile log-likelihood of $\beta$ derived from (\ref{eq:ET-lik}) in the exponential tilt model based on the labeled data only
or equivalently the log-likelihood of $(\beta_0^c,\beta_1)$ as (\ref{eq:logistic-lik}) from logistic regression based on the labeled data only, after the intercept shift $\beta_0^c= \beta_0 + \log(n_2/n_1)$.
We show that the MLE $\hat\beta$ from ETM model (\ref{eq:ETM1})--(\ref{fm:exp-tilt2}) or equivalently SLR model (\ref{eq:new-logit1})--(\ref{eq:new-logit3})
is asymptotically more efficient than that from logistic regression based on the labeled data only.

\begin{proposition} \label{prop:efficiency}
Denote by $\hat \beta^\ell$ the estimator of $\beta$ obtained by maximizing $\kappa\{ \rho^\ell, \beta, \tilde\alpha(\rho^\ell)\}$
or equivalently by logistic regression based on the labeled data only.
Then the asymptotic variance matrix of the MLE $\hat\beta$ from  ETM model (\ref{eq:ETM1})--(\ref{fm:exp-tilt2}) 
is no greater (in the usual order on positive-definite matrices) than
that of $\hat\beta^\ell$ under standard regularity conditions.
\end{proposition}

\subsection{Regularized estimation and EM algorithm} \label{sec:EM}

The results in Section~\ref{sec:SLR} provide theoretical support for the use of the objective functions $\kappa \{ \rho, \beta, \hat\alpha(\beta)\}$ and $\kappa\{ \rho, \beta, \tilde\alpha(\rho)\}$.
In real applications, the MLE $(\hat\rho,\hat\beta)$ may not behave satisfactorily as predicted by standard asymptotic theory for various reasons.
The labeled sample size may not be sufficiently large. The dimension of the feature vector or the complexity of functions of features may be too high, compared with the labeled and unlabel data sizes.
Therefore, we propose regularized estimation by adding suitable penalties to the objective functions.

For the coefficient vector $\beta_1$, we employ a ridge penalty $\lambda \|\beta_1\|_2^2$, although alternative penalties can also be allowed including a Lasso penalty.
For the mixture proportion $\rho$, we use a penalty in the form of the log density of a Beta distribution, $\tau_1 \log(1-\rho) + \tau_2\log\rho$,
where $\tau_1 = \gamma (1-\rho^0)  n_3/N$ and $\tau_2 = \gamma \rho^0  n_3/N$ for a ``center'' $\rho^0 \in (0,1)$ and a ``scale'' $\gamma \ge 0$. This choice is
motivated by conceptual and computational simplicity in the EM algorithm to be discussed.
Combining these penalties with $\kappa \{ \rho, \beta, \hat\alpha(\beta)\}$ gives the following penalized objective function
\begin{align}
\kappa \{ \rho, \beta, \hat\alpha(\beta)\}  -  \lambda \|\beta_1\|_2^2  +  \gamma (1-\rho^0)  (n_3/N) \log(1-\rho) + \gamma \rho^0  (n_3/N) \log\rho. \label{eq:profile-obj}
\end{align}
Similarly, the penalized objective function based on $\kappa\{ \rho, \beta, \tilde\alpha(\rho)\}$ is
\begin{align}
\kappa\{ \rho, \beta, \tilde\alpha(\rho)\}  -  \lambda \|\beta_1\|_2^2  +  \gamma (1-\rho^0)  (n_3/N) \log(1-\rho) + \gamma \rho^0  (n_3/N) \log\rho. \label{eq:approx-obj}
\end{align}
Maximization of (\ref{eq:profile-obj}) or (\ref{eq:approx-obj}) will be called profile or direct SLR respectively.
The two methods in general lead to different estimates of $(\rho,\beta)$ when $\gamma >0$, although they can be shown to be equivalent similarly as in Proposition~\ref{prop:equiv}
when $\gamma=0$. In fact, as $\gamma \to \infty$ (i.e., $\rho$ is fixed as $\rho^0$), the estimator of $\beta$ from profile SLR is known to
asymptotically more efficient than from direct SLR (Tan 2009).

We construct EM algorithms (Dempster et al.~1977) to numerically maximize (\ref{eq:profile-obj}) and (\ref{eq:approx-obj}).
Of particular interest is that these algorithms shed light on the effect of the regularization introduced.
Various other optimization techniques can also be exploited, because $\kappa\{ \rho, \beta, \tilde\alpha(\rho)\}$ is directly of a closed form,
and $\kappa\{ \rho, \beta, \hat\alpha(\beta)\}$ is defined only after univariate minimization in $\alpha$.


We describe some details about the EM algorithm for profile SLR.\
See the Supplement (Section~\ref{sec:dSLR-EM}) for the corresponding algorithm for direct SLR.
We return to the nonparametric log-likelihood (\ref{eq:full-lik}) and introduce the following data augmentation.
For $j=1,2,3$, let $u_{ji} \sim \mbox{Bernoulli}\,(\rho_j)$ such that
$ (x_{ji} | u_{ji}=0 )\sim G_0$ and $(x_{ji} | u_{ji}=1 )\sim G_1$.
Recall that $\rho_1=0$ and $\rho_2=1$ and hence $u_{1i}=0$ and $u_{2i} = 1$ fixed.
Denote the penalty term in (\ref{eq:profile-obj}) or (\ref{eq:approx-obj}) as $\mbox{pen}(\rho,\beta)$.

{\bf E-step.} The expectation of the augmented objective given the current estimates $(\rho^{(t)},\beta^{(t)})$ is\vspace{-.1in}
\begin{align}
Q^{(t)}(\rho,\beta, G_0) & =\frac{1}{N} \sumjminj \Big[ (1-\E^{(t)}  u_{ji})\log \{ (1-\rho_j)G_0(\{x_{ji}\})\}  \nonumber \\
& \quad + \E^{(t)} u_{ji}\log \{\rho_j \exp(\beta_0+ \beta_1^\T x_{ji}) G_0(\{x_{ji}\}) \} \Big] + \mbox{pen}(\rho,\beta), \label{eq:EM-Q}
\end{align}
where $\E^{(t)}  u_{ji} = \rho^{(t)}_j \exp(\beta^{(t)}_0+ \beta_1^{(t)\T} x_{ji}) / \{1- \rho^{(t)}_j +\rho^{(t)}_j  \exp(\beta^{(t)}_0+ \beta_1^{(t)\T} x_{ji}) \}$.

{\bf M-step.} The next estimates $(\rho^{(t+1)}, \beta^{(t+1)})$ are obtained as a maximizer of the expected objective (\ref{eq:EM-Q}) with $G_0$ profiled out, that is,
$\pQ^{(t)} (\rho, \beta) = \max_{G_0} Q^{(t)}(\rho,\beta, G_0)$ over all possible $G_0$ which is a probability measure supported on the pooled data $\{x_{ji} : i=1,\ldots,n_j, j=1,2,3\}$ with
$\int \exp(\beta_0 + \beta_1^\T x) \dd G_0 =1$. In correspondence to $\kappa(\rho,\beta,\alpha)$, denote\vspace{-.1in}
\begin{align*}
\kappa_Q^{(t)}  (\rho,\beta,\alpha) & = \frac{1}{N} \sumjminj \Big[ (1-\E^{(t)}  u_{ji})\log  \left\{\frac{1-\rho_j}{1-\alpha+ \alpha \exp(\beta_0+\beta_1^\T x_{ji})} \right\} \\
& \quad  + \E^{(t)} u_{ji} \log \left\{\frac{\rho_j \exp(\beta_0+ \beta_1^\T x_{ji}) }{ 1-\alpha+ \alpha \exp(\beta_0+\beta_1^\T x_{ji}) }  \right\}\Big] - \log N + \mbox{pen}(\rho,\beta) .
\end{align*}
Instead of maximizing $\pQ^{(t)} (\rho, \beta)$ directly, we find a simple scheme for computing $(\rho^{(t+1)}, \beta^{(t+1)})$.

\begin{proposition} \label{prop:EM-formula}
Let\vspace{-.1in}
\begin{align}
\rho^{(t+1)} = \frac{ n_3^{-1} \sum_{i=1}^{n_3} \E^{(t)}  u_{ji}  + \gamma \rho^0}{1+\gamma} ,\quad
\alpha^{(t+1)} = \frac{1}{N} \sumjminj  \E^{(t)}  u_{ji} . \label{eq:EM-update}
\end{align}
If and only if $\beta^{(t+1)}$ is a local (or global) maximizer of  $\kappa_Q^{(t)}  (\rho^{(t+1)},\beta,\alpha^{(t+1)})$,
then $(\rho^{(t+1)},\beta^{(t+1)})$ is a local (or respectively global) maximizer of  $\pQ^{(t)} (\rho, \beta)$.
\end{proposition}

Proposition~\ref{prop:EM-formula} is useful both computationally and conceptually. First, $\rho^{(t+1)}$ is of a closed form,
as a weighted average, with the weight depending on the scale $\gamma$, between
the prior center $\rho^0$ and the empirical estimate $n_3^{-1} \sum_{i=1}^{n_3} \E^{(t)}  u_{ji}$, which would be obtained  with $\gamma\to\infty$ or $\gamma=0$ respectively.
Moreover, $\beta^{(t+1)}$ can be equivalently computed by maximizing the objective function\vspace{-.1in}
\begin{align}
 \hspace*{-.05in} \frac{1}{N} \sumjminj \Big[ \E^{(t)}  u_{ji} (\beta_0+ \beta_1^\T x_{ji}) - \log\{1-\alpha^{(t+1)}+ \alpha^{(t+1)} \exp(\beta_0+\beta_1^\T x_{ji})\} \Big]
 - \lambda \|\beta_1\|_2^2 , \label{eq:EM-beta-obj}
\end{align}
which is concave in $\beta$ and of a similar form to the log-likelihood (\ref{eq:logistic-lik}) with a ridge penalty for logistic regression.
Each imputed probability $\E^{(t)}  u_{ji}$ serves as a pseudo response.

In our implementation, the prior center $\rho^0$ is fixed as $\rho^\ell= n_2/n$, the proportion of $y=1$ in the labeled sample, and
the scales $(\lambda,\gamma)$ are treated as tuning parameters, to be selected by cross validation.
Numerically, this procedure allows an adaptive interpolation between the two extremes: a fixed choice $\rho^\ell$
or an empirical estimate by maximum likelihood.
For direct SLR (but not profile SLR), our adaptive procedure reduces to and hence accommodates logistic regression with labeled data only at one extreme with $\gamma\to\infty$.
See the Supplement (Section~\ref{sec:dSLR-EM}) for further discussion.

\vspace{-.05in}
\section{Related work}
\vspace{-.05in}

There is a vast literature on semi-supervised learning. See, for example, Chapelle et al.~(2006) and Zhu (2008). For space limitation, we only discuss
directly related work to ours.

{\bf Generative models and EM.} A generative model can be postulated for $(y,x)$ jointly such that
$p(y,x; \rho,\theta) = p(y; \rho) p(x| y; \theta)$, where $\rho$ denotes the label proportion and $\theta$ denotes the parameters
associated with the feature distributions given labels (e.g., Nigam et al.~2000).
In our notation, a generative model corresponds to Eqs~(\ref{eq:ETM1})--(\ref{eq:ETM3}), but with both $G_0$ and $G_1$ parametrically specified.
For training by EM algorithms,
the expected objective in the E-step is similar to $Q^{(t)}(\rho,\beta, G_0)$ in (\ref{eq:EM-Q}), except that $G_k(\{x_{ji}\})$
is replaced by $p(x_{ji} | y_{ji}=k; \theta)$ for $k=0$ or 1.
The performance of  generative modeling can be sensitive to whether the model assumptions are correct or not (Cozman et al.~2003).  
In this regard, our approach based on ETM models is attractive in only specifying a parametric form (\ref{fm:exp-tilt2}) for the density ratio between $G_0$ and $G_1$
while leaving the distribution $G_0$ nonparametric.

{\bf Logistic regression and EM.} There are various efforts to extend logistic regression in an EM-style for semi-supervised learning.
Notably, Amini \& Gallinari (2002) proposed a classification EM algorithm using logistic regression (\ref{fm:logistic_regression}),
which can be described as follows: \vspace{-.08in}
\begin{itemize} \addtolength{\itemsep}{-.05in}
\item E-step: Compute $\E^{(t)}  u_{3i} = \{1+ \exp(-\beta^{c(t)}_0-\beta_1^{(t)\T} x_{3i}) \}$. Fix $\E^{(t)} u_{1i}=0$ and $\E^{(t)} u_{2i}=1$.

\item C-step: Let $u^{(t)}_{3i}=1$ if  $\E^{(t)}  u_{3i} \ge .5$ and 0 otherwise. Fix $u^{(t)}_{1i}=0$ and $u^{(t)}_{2i}=1$.

\item M-step: Compute $(\beta^{c(t+1)}_0, \beta_1^{(t+1)} )$ by maximizing the objective
$\sumjminj [ u^{(t)}_{ji} (\beta^c_0+ \beta_1^\T x_{ji}) - \log\{1+ \exp(\beta^c_0+\beta_1^\T x_{ji})\} ]$.
\end{itemize}\vspace{-.08in}
Although convergence of classification EM was studied for clustering (Celeux \& Govaert 1992), it seems unclear what objective function is optimized by the preceding algorithm.
A worrisome phenomenon we notice is that if soft classification is used instead of hard classification, then the algorithm merely optimizes the log-likelihood of logistic regression with the labeled data only.
By comparing (\ref{eq:EM-beta-obj}) and (\ref{eq:CEM-obj2}),
this modified algorithm can be shown to reduce to our EM algorithm with $\rho^{(t)}$ and $\alpha^{(t)}$ clamped at $\rho^\ell=n_2/n$, the proportion of $y=1$ in the labeled sample.

\begin{proposition} \label{prop:CEM}
If the objective in the M-step is modified with $u^{(t)}_{ji}$ replaced by $\E^{(t)} u_{ji}$ as \vspace{-.1in}
\begin{align}
\sumjminj \Big[ \E^{(t)} u_{ji} (\beta^c_0+ \beta_1^\T x_{ji}) - \log\{1+ \exp(\beta^c_0+\beta_1^\T x_{ji})\} \Big] . \label{eq:CEM-obj2}
\end{align}

\vspace*{-.2in}
then $(\beta^{c(t)}_0, \beta_1^{(t)})$ converges as $t\to\infty$ to MLE of logistic regression based on the labeled data only.
\end{proposition}

We notice that the conclusion also holds if (\ref{eq:CEM-obj2}) is replaced by the cost function proposed in Wang et al.~(2009), Eq~(2), when the logistic loss is used as the cost function
on labeled data.

{\bf Regularized methods.} Various methods have been proposed by introducing a regularizer depending on unlabeled data to the
log-likelihood of logistic regression with labeled data. Examples include entropy regularization (Grandvalet \& Bengio 2005),
expectation regularization (Mann \& McCallum 2007), and graph-based priors (Krishnapuram et al.~2005).
An important difference from our methods is that these penalized objective functions seem to be Fisher consistent only when they
reduce to the log-likelihood of logistic regression with labeled data only alone, regardless of unlabeled data.
For another difference, the class proportions in unlabeled data are implicitly assumed to be the same as in labeled data in entropy regularization,
and need to be explicitly estimated from labeled data or external knowledge in the case of label regularization (Mann \& McCallum 2007).

\vspace{-.05in}
\section{Numerical experiments} \label{sec:experiment}
\vspace{-.05in}

We report experiments on 15 benchmark datasets including 11 UCI datasets and 4 SSL benchmark datasets.
We compare our methods, profile SLR (pSLR) and direct SLR (dSLR), with
2 supervised methods, ridge logistic regression (RLR) and SVM, and 2 semi-supervised methods, entropy regularization (ER) (Grandvalet \& Bengio 2005) and transductive SVM (TSVM) (Joachims 1999).
For each method, only linear predictors are studied. All tuning parameters are selected by 5-fold cross validation.
See the Supplement (Section~\ref{sec:experiment-details}) for details about the datasets and implementations.

For each dataset except SPAM, a training set is obtained as follows: labeled data are sampled for a certain size (25 or 100) and fixed class proportions and then unlabeled data
are sampled such that the labeled and unlabeled data combined are 2/3 of the original dataset. The remaining 1/3 of the dataset is used as a test set.
For SPAM, the preceding procedure is applied to a subsample of size 750 from the original dataset.
To allow different class proportions between labeled and unlabeled data, we consider two schemes: the class proportions in the labeled data are close to those of the original dataset (``Homo Prop''),
or larger (or smaller) than the latter by an odds ratio of 4 (``Flip Prop'') if the odds of positive versus negative labels is $\le 1$ (or respectively $> 1$) in the original dataset.
Hence the class balance constraint as used in TSVM is misspecified in the second scheme.

\begin{table}[t]
\caption{Classification accuracy in \% (mean $\pm$ sd) on test data over 20 repeated runs, with labeled training data size 100. Subscript $_a$ indicates that intercept adjustment is applied (see the text).}\label{Table:labeled-100-acc} \vskip -.05in
\resizebox{\columnwidth}{!}{%
\begin{tabular}{ccccccc}  \toprule
Homo Prop  & RLR           & ER                     & pSLR                   & dSLR                   & SVM                    & TSVM                   \\\midrule
AUSTRA   & 85.37 $\pm$ 2.00 & \textbf{85.50 $\pm$ 1.94} & 85.43 $\pm$ 2.07          & 85.33 $\pm$ 2.03          & 85.37 $\pm$ 1.96 & 85.15 $\pm$ 1.79          \\
BCW            & 95.76 $\pm$ 1.04 & 95.64 $\pm$ 1.08          & 95.71 $\pm$ 1.07          & 95.80 $\pm$ 1.07          & 96.13 $\pm$ 1.04          & \textbf{96.44 $\pm$ 0.92} \\
GERMAN         & 72.16 $\pm$ 2.69 & 72.22 $\pm$ 2.60          & \textbf{72.39 $\pm$ 2.71} & 72.12 $\pm$ 2.68          & 70.65 $\pm$ 2.85          & 69.01 $\pm$ 3.94          \\
HEART          & 80.94 $\pm$ 3.56 & 81.04 $\pm$ 4.23          & \textbf{81.46 $\pm$ 3.41} & 80.94 $\pm$ 3.78          & 80.00 $\pm$ 4.35          & 80.36 $\pm$ 5.00          \\
INON           & 84.38 $\pm$ 2.29 & 84.17 $\pm$ 2.16          & \textbf{85.00 $\pm$ 1.21} & 84.38 $\pm$ 2.29          & 83.92 $\pm$ 2.51          & 83.33 $\pm$ 3.18          \\
LIVER & 65.57 $\pm$ 4.37 & 65.83 $\pm$ 4.20          & 66.35 $\pm$ 4.65          & 66.22 $\pm$ 4.22          & \textbf{67.87 $\pm$ 3.22} & 64.83 $\pm$ 5.00          \\
PIMA           & 74.71 $\pm$ 2.99 & \textbf{75.06 $\pm$ 3.07} & 75.02 $\pm$ 2.90          & 74.71 $\pm$ 3.04          & 74.65 $\pm$ 2.69          & 72.58 $\pm$ 3.97          \\
SPAM           & 87.98 $\pm$ 2.70 & \textbf{88.04 $\pm$ 2.95} & 87.90 $\pm$ 2.49          & 87.94 $\pm$ 2.70          & 85.74  $\pm$ 3.90          & 87.04 $\pm$ 5.37          \\
VEHICLE        & 93.10 $\pm$ 2.73 & 92.59 $\pm$ 2.80          & 92.45 $\pm$ 2.83          & \textbf{93.24 $\pm$ 2.82} & 92.38 $\pm$ 3.31          & 93.03 $\pm$ 3.39          \\
VOTES          & 93.66 $\pm$ 2.55 & 93.59 $\pm$ 2.59          & 93.66 $\pm$ 2.32          & 93.45 $\pm$ 2.59          & \textbf{94.17 $\pm$ 2.57} & 94.03 $\pm$ 2.92          \\
WDBC           & 95.92 $\pm$ 1.65 & 95.61 $\pm$ 1.70          & 95.89 $\pm$ 1.48          & 95.92 $\pm$ 1.65          & 95.67 $\pm$ 1.55          & \textbf{96.06 $\pm$ 1.31} \\
BCI            & 66.50 $\pm$ 4.06 & 65.83 $\pm$ 3.80          & 65.86 $\pm$ 4.89          & 65.86 $\pm$ 4.40          & \textbf{68.46 $\pm$ 5.01} & 67.48 $\pm$ 5.20          \\
COIL           & 78.95 $\pm$ 3.15 & 78.96 $\pm$ 3.24          & 79.07 $\pm$ 3.89          & 78.70 $\pm$ 3.42          & 80.10$\pm$2.53          & \textbf{81.39 $\pm$ 2.38} \\
DIGIT1         & 89.90 $\pm$ 1.11 & 89.29 $\pm$ 2.70          & \textbf{90.00 $\pm$ 1.18} & 89.87 $\pm$ 1.16          & 89.30$\pm$1.33          & 89.73 $\pm$ 1.45          \\
USPS           & 85.39 $\pm$ 2.38 & 85.62 $\pm$ 2.26          & \textbf{85.97 $\pm$ 1.91} & 85.54 $\pm$ 2.45          & 85.50$\pm$2.13        & {84.71 $\pm$ 2.27} \\\midrule
Average accuracy &83.35	&	83.27&	\textbf{83.48}&	83.33	&83.34&	83.01\\ 
 \# within 1\% of highest&      \textbf{12/15}&	\textbf{12/15}&	\textbf{12/15}&	\textbf{12/15}&	10/15&	7/15\\
\bottomrule
\end{tabular}
}
\noindent

\resizebox{\columnwidth}{!}{%
\begin{tabular}{ccccccc}\toprule
 Flip Prop & RLR$_a$     & ER$_a$                   & pSLR$_a$             & dSLR$_a$            & SVM$_a$                  & TSVM          \\\midrule
AUSTRA     & 84.98 $\pm$ 2.18 & 85.00 $\pm$ 2.43          & 85.07 $\pm$ 2.86          & 85.46 $\pm$ 2.35          & \textbf{85.11$\pm$2.32} & 71.78 $\pm$ 7.05 \\
BCW            & 96.27 $\pm$ 1.39 & 96.07 $\pm$ 1.50          &\textbf{ 96.64 $\pm$ 1.46   }       & 96.47 $\pm$ 1.36          & 96.00$\pm$1.76          & 95.80 $\pm$ 2.80  \\
GERMAN         & 68.77 $\pm$ 2.36 & 68.35 $\pm$ 2.27          & 68.05 $\pm$ 2.42          & \textbf{69.61 $\pm$ 2.25} & 68.30$\pm$2.27          & 57.57 $\pm$ 4.00 \\
HEART          & 80.36 $\pm$ 4.75 & 80.10 $\pm$ 4.62          & 81.82 $\pm$ 3.60          & \textbf{81.98 $\pm$ 3.61} & 79.06$\pm$4.73          & 62.71 $\pm$ 4.34 \\
INON           &\textbf{ 83.96 $\pm$ 3.52} & 82.75 $\pm$ 3.90          & 82.81 $\pm$ 3.94          & 83.75 $\pm$ 2.64 & 80.96$\pm$5.02          & 59.22 $\pm$ 8.40 \\
LIVER          & 60.70 $\pm$ 6.60 & 60.30 $\pm$ 6.71          & 62.00 $\pm$ 6.91          & \textbf{62.83 $\pm$ 5.80} & 59.70$\pm$8.82          & 54.30 $\pm$ 2.19 \\
PIMA           & 71.39 $\pm$ 3.53 & 71.50 $\pm$ 3.38          & \textbf{72.03 $\pm$ 2.95} & 71.84 $\pm$ 3.28          & 71.66$\pm$3.28          & 61.80 $\pm$ 2.86 \\
SPAM           & 88.52 $\pm$ 2.38 & \textbf{89.06 $\pm$ 2.97} & 88.60 $\pm$ 2.75          & 88.44 $\pm$ 2.34          & 86.68$\pm$4.28          & 87.22 $\pm$ 5.30 \\
VEHICLE        & 89.93 $\pm$ 6.31 & 88.79 $\pm$ 5.70          & 91.55 $\pm$ 5.55          & \textbf{93.34 $\pm$ 1.97} & 88.66$\pm$4.02          & 70.62 $\pm$ 4.63 \\
VOTES          & 92.72 $\pm$ 2.04 & 92.66 $\pm$ 2.25          & 93.31 $\pm$ 1.55          & \textbf{93.31 $\pm$ 1.58} & 92.72$\pm$2.92          & 81.03 $\pm$ 6.02 \\
WDBC           & 96.20 $\pm$ 1.59 & 96.33 $\pm$ 1.40          & 96.71 $\pm$ 1.58          & \textbf{97.22 $\pm$ 1.68} & 95.97$\pm$1.43          & 80.23 $\pm$ 4.81 \\
BCI            & 62.78 $\pm$ 3.81 & 62.71 $\pm$ 4.10          & 65.86 $\pm$ 5.13          & \textbf{66.35 $\pm$ 4.91} & 65.04$\pm$4.21          & 60.83 $\pm$ 3.83 \\
COIL           & 71.86 $\pm$ 6.59 & 72.36 $\pm$ 6.93          & 72.20 $\pm$ 7.72          & \textbf{73.45 $\pm$ 6.56} & 69.01$\pm$9.81         & 66.28 $\pm$ 2.90 \\
DIGIT1         & 87.93 $\pm$ 2.11 & 87.08 $\pm$ 4.38          & 87.68 $\pm$ 2.60          & \textbf{88.89 $\pm$ 2.73} & 87.16$\pm$2.39          & 73.10 $\pm$ 2.03 \\
USPS           & 82.05 $\pm$ 3.42 & 82.29 $\pm$ 3.29          & \textbf{83.74 $\pm$ 3.21} & 83.65 $\pm$ 3.17          & 81.16$\pm$3.06          & 64.95 $\pm$ 1.90\\\midrule
Average accuracy & 81.23& 	 81.02	& 81.87& 	\textbf{82.44}& 	80.48	& 69.83\\ 
 \# within 1\% of highest&   8/15&	6/15&	10/15&	\textbf{15/15}&	4/15	&1/15\\\bottomrule
\end{tabular}
}  \vskip -.15in
\end{table}

Care is needed to define classifiers on test data. In the Homo Prop scheme, the 4 existing methods are applied as usual,
and accordingly the classifiers from our methods are the sign of $\log(n_2/n_1) + \hat\beta_0 + \hat\beta_1^\T x$, where $(n_1,n_2)$ are the class sizes in the labeled training data.
In the Flip Prop scheme, the classifiers from RLR, LR, and SVM are the sign of $-\log(n_2/n_1)+\tilde\beta^c_0 + \tilde\beta_1^\T x$,
and those from our methods are the sign of $\hat\beta_0 + \hat\beta_1^\T x$. Hence the intercepts of linear predictors are adjusted by assuming 1:1
class proportions in the test data. This assumption is often invalid in our experiments, but seems neutral when the actual class proportions in test data are unknown.
The ``linear predictor'' is converted by logit from class probabilities for SVM, but this is currently unavailable for TSVM.
Alternatively, class weights can be used in SVM, but this technique has not been developed for TSVM.

Table \ref{Table:labeled-100-acc} presents the results with labeled data size 100. See the Supplement for those with labeled data size 25 and AUC results.
In the Homo Prop scheme, the logistic-type methods, RLR, ER, pSLR, and dSLR, perform similarly to each other, and noticeably better than SVM and TSVM in terms of
accuracy achieved within 1\% of the highest (in bold). While unstable performances of SVM and TSVM have been previously noticed (e.g., Li \& Zhou 2015),
such good performances of RLR and ER on these benchmark datasets appear not to have been reported before.
In the Flip Prop scheme, our methods, dSLR and pSLR, achieve the best two performances, sometimes with considerable margins of improvement over other methods.
In this case, all methods except TSVM are applied with intercept adjustment as described above.
Because which proportion scheme holds may be unknown in practice, the results with intercept adjustment in the Homo Prop scheme are reported in the Supplement.
Our methods remain to achieve close to the best performance among the methods studied.

\vspace{-.08in}
\section{Conclusion} \label{sec:conclusion}
\vspace{-.1in}

We develop an extension of logistic regression for semi-supervised learning, with strong support from statistical theory, algorithms, and numerical results.
There are various questions of interest for future work. Our approach can be readily extended by employing nonlinear predictors such as
kernel representations or neural networks. Further experiments with such extensions are desired, as well as applications to more complex text and image classification.

\newpage
\section*{References}


\small

Amini, M.R.  \& Gallinari, P.  \  (2002)  Semi-supervised logistic regression. \textit{Proceedings of the 15th European Conference on Artificial Intelligence}, 390--394.

Bartlett, P., Jordan, M., \& McAuliffe, J. \ (2006) Convexity, classification, and risk bounds. \textit{Journal of the American Statistical Association}, 101, 138--156.

Celeux, G. \& Govaert, G. \ (1992) A classification EM algorithm and two stochastic versions.  \textit{Computational Statistics and Data Analysis}, 14,  315--332.

Chapelle, O., Zien, A. \& Sch\"olkopf, B. \ (2006) \textit{Semi-Supervised Learning}. MIT Press.

Cozman, F., Cohen, I. \& Cirelo, M. \ (2003). Semi-supervised learning of mixture
models.  \textit{Proceedings of the 20th International Conference on Machine Learning},  99--106.

Dempster, A.P., Laird, N.M. \& Rubin, D.B. (1977) Maximum likelihood from incomplete data via the EM algorithm. \textit{Journal of the Royal Statistical Society, Series B}, 39, 1--22.

Grandvalet, Y.,  \& Bengio, Y.  \ (2005) Semi-supervised learning by entropy minimization. \textit{Advances in Neural Information Processing Systems 17},  529--536.


Joachims, T.\ (1999) Transductive inference for text classification using support vector machines. \textit{Proceedings of the 16th International Conference on Machine Learning},  200--209.

Kiefer, J. \& Wolfowitz, J. \  (1956) Consistency of the maximum likelihood estimator in the presence of infinitely many incidental parameters. \textit{Annals of Statistics}, 27, 887--906.

Krishnapuram, B., Williams, D., Xue, Y., Carin, L., Figueiredo, M. \& Hartemink, A.J. \ (2005) On semi-supervised classification. \textit{Advances in Neural Information Processing Systems 17},  721--728.


Li, Y.-F. \& Zhou, Z.-H. \ (2015) Towards making unlabeled data never hurt. \textit{IEEE Transactions on Pattern analysis and Machine Intelligence}, 37,  175--188.

Lin, Y. (2002) Support vector machines and the Bayes rule in classification. \textit{Data Mining and Knowledge Discovery}, 6, 259--275.

Mann, G.S. \& McCallum, A. \ (2007) Simple, robust, scalable semi-supervised learning via expectation regularization. \textit{Proceedings of the 24th International Conference on Machine learning},  593--600.


Manski, C.F. \ (1988) \textit{Analog Estimation Methods in Econometrics}, Chapman \& Hall.

Nigam, K., McCallum, A.K., Thrun, S. \& Mitchell, T. (2000) Text classification from labeled and unlabeled documents using EM. \textit{Machine learning}, 39, 103--134.

Owen, A.B. \ (2001) \textit{Empirical Likelihood}. Chapman \& Hall/CRC.

Prentice, R.L. \& Pyke, R. \ (1979) Logistic disease incidence models and case-control studies. \textit{Biometrika}, 66,  403--411.

Qin, J. \ (1998) Inferences for case-control and semiparametric two-sample density ratio models. \textit{Biometrika}, 85, 619--630.

Qin, J. \ (1999) Empirical likelihood ratio based confidence intervals for mixture proportions. \textit{Annals of Statistics}, 27,  1368--1384.

Tan, Z. \  (2009) A note on profile likelihood for exponential tilt mixture models. \textit{Biometrika}, 96, 229--236.

Wang, J., Shen, X. \& Pan, W. \ (2009) On efficient large margin semisupervised learning: Method and theory. \textit{Journal of Machine Learning Research}, 10, 719--742.

Vapnik, V. \ (1998) \textit{Statistical Learning Theory}. Wiley-Interscience.

White, H. \ (1982) Maximum likelihood estimation of misspecified models. \textit{Econometrica}, 50, 1--25.


Zhu, X.J. \ (2008) Semi-supervised learning literature survey. Technical Report, University of Wisconsin-Madison, Department of Computer Sciences.

Zou, F., Fine, J.P. \& Yandell, B.S. \ (2002) On empirical likelihood for a semiparametric mixture model. \textit{Biometrika}, 89, 61--75.

\normalsize

\clearpage

\setcounter{page}{1}

\setcounter{section}{0}
\setcounter{equation}{0}

\renewcommand{\thesection}{\Roman{section}}
\renewcommand{\theequation}{S\arabic{equation}}

\setcounter{figure}{0}
\setcounter{table}{0}

\renewcommand\thefigure{S\arabic{figure}}
\renewcommand\thetable{S\arabic{table}}

\setcounter{proposition}{0}
\renewcommand{\theproposition}{S\arabic{proposition}}

\setcounter{lemma}{0}
\renewcommand\thelemma{S\arabic{lemma}}

\setcounter{footnote}{0}

  \hrule height 4pt
  \vskip 0.25in

  \begin{center}
  {\LARGE\bf Supplementary Material for\\[.05in]
    ``Semi-supervised Logistic Learning Based on Exponential Tilt Mixture Models''}
  \end{center}

  \vskip 0.29in
  \hrule height 1pt
  \vskip 0.09in%

  \vspace{.1in}
   \begin{center}
   \begin{tabular}[t]{c}
   {\bf Xinwei Zhang \, \& \, Zhiqiang Tan} \\
   Department of Statistics, Rutgers University, USA \\
   \texttt{xinwei.zhang@rutgers.edu, ztan@stat.rutgers.edu} \\
   \end{tabular}%
   \end{center}

\section{Introduction}

We provide additional material to support the content of the paper.
All equation and proposition numbers referred to are from the paper, except S1, S2, etc.

\section{Illustration} \label{sec:illustration}

\begin{figure}[!hb]
  \centering
  \rule[-.5cm]{4cm}{0cm}
   \includegraphics[width=\textwidth, height=2in]{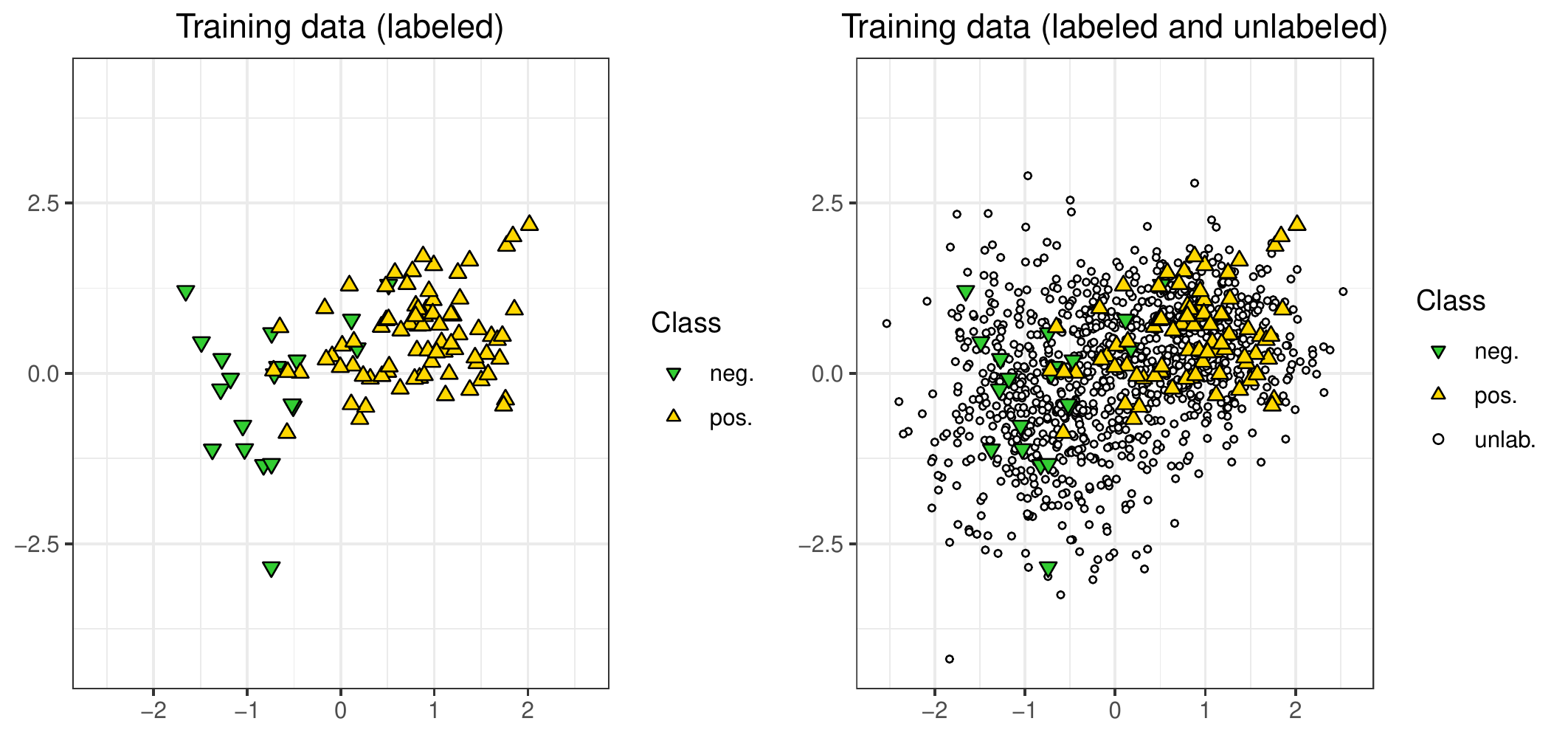}\rule[-.5cm]{4cm}{0cm}
  \caption{Training data from bivariate Gaussian} \label{Fig:illustrate_data}
\end{figure}

We provide a simple example to highlight comparison between new and existing methods.
A labeled sample of size 100 is drawn, where 20 are from bivariate Gaussian, $G_0$, with mean $(-6,-6)$ and diagonal variance matrix $(5^2, 15^2)$,
and 80 are from bivariate Gaussian, $G_1$, with mean $(6,6)$ and diagonal variance matrix $(5^2, 10^2)$. An unlabeled sample of size 1000
is drawn, where $500$ are from $G_0$ and $500$ from $G_1$ and then the labels are removed.
This is similar to the Flip Prop scheme in numerical experiments in Section~\ref{sec:experiment}, where the class proportions in unlabeled data differ from those in labeled data.
The training set including both labeled and unlabeled data is then rescaled such that the root mean square of each feature is 1, as shown in Figure \ref{Fig:illustrate_data}.

\begin{figure}[!ht]
  \centering
  \rule[-.5cm]{4cm}{0cm}
   \includegraphics[width=\textwidth, height=8in]{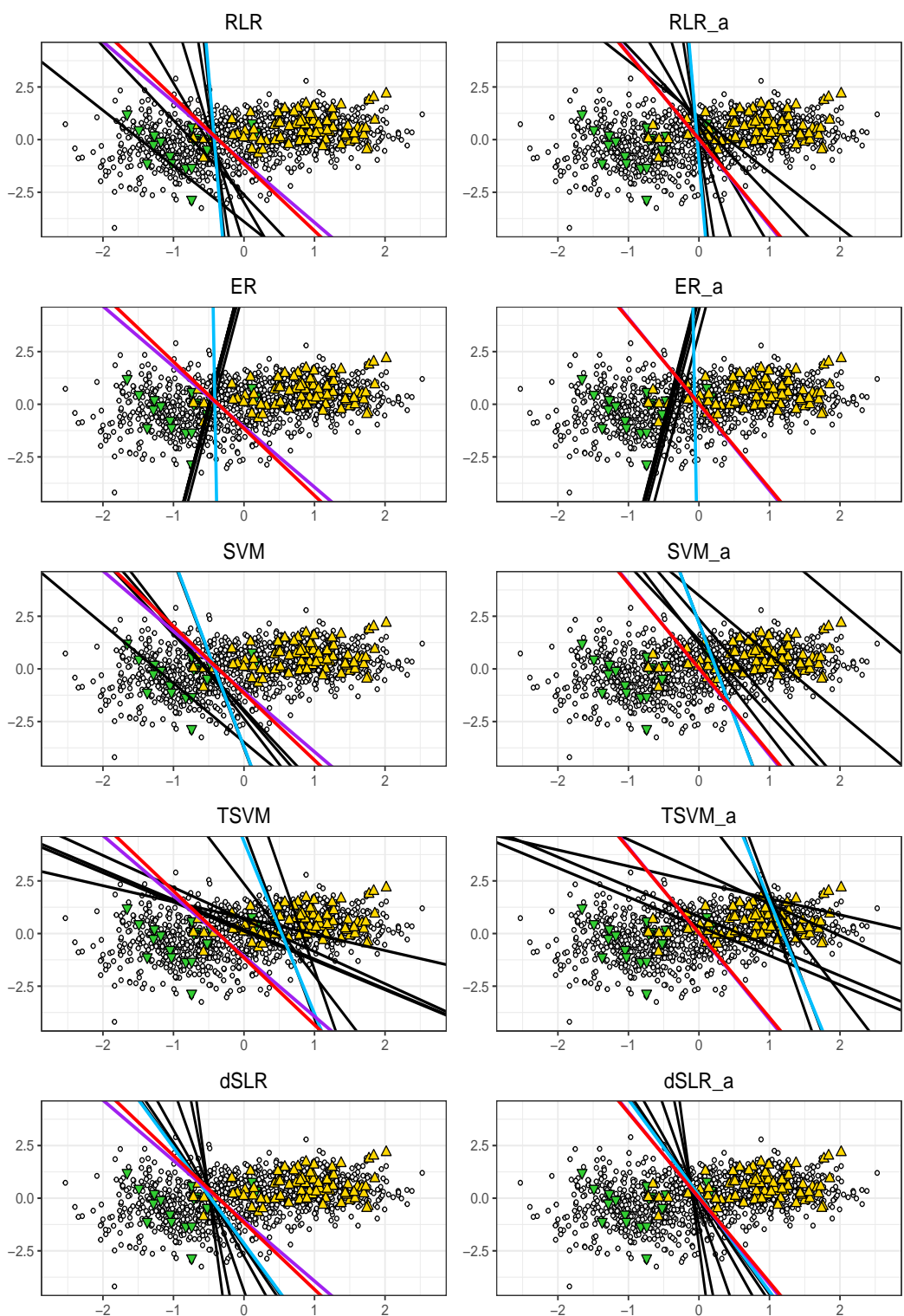}\rule[-.5cm]{4cm}{0cm}
  \caption{Decision lines with bivariate Gaussian data} \label{Fig:illustrate}
\end{figure}

Figure \ref{Fig:illustrate} (rows 1 to 5) shows the decision lines from, respectively, ridge linear regression (RLR), entropy regularization (ER), SVM, TSVM, and direct SLR (dSLR).
In the left column are the decision lines without intercept adjustment (corresponding to an assumption of 1:4 class proportions in test data as in labeled training data),
and in the right column are those with intercept adjustment
(corresponding to an assumption of 1:1 class proportions in test data as in unlabeled training data), as described in Section~\ref{sec:experiment}.
In practice, the class proportions in test data may be unknown and hence some assumption is needed.\footnote{The assumption of 1:1 class proportions in test data is used
to define classifiers in the Flip Prop scheme in Section~\ref{sec:experiment}, even though this assumption is violated for a majority of datasets studied (see Table~\ref{Table:data_sets_stat}).}
For ease of comparison, the intercept adjustment is directly applied to $w^\T x + b$ for SVM and TSVM, instead of the ``linear predictor'' converted by logit from class probabilities (if available),
which would yield a nonlinear decision boundary.
Alternatively, class weights can be used in SVM to account for differences in class proportions between training and test data. But this technique has not been developed for TSVM.

For each method, eight decision lines (black or blue) are plotted, by using 8 values of a tuning parameter. Some of the lines may fall outside the plot region.
The blue lines correspond to the least amount of penalization used, that is, smallest $\lambda$, $\lambda_e$ and $\gamma$ and largest $C$.
See Section~\ref{sec:experiment-details} later for a description of the tuning parameters involved.
For RLR, $\log_{10}(\lambda)$ is varied uniformly from $[-5,-1]$.
For ER, $\lambda_e$ is varied uniformly from $0.01$ from 1, while $\lambda$ is fixed at 0, to isolate the effect of entropy regularization.
For SVM and TSVM, $\log_{10}(C)$ is varied uniformly from  $[-2, 2]$. For TSVM, the parameter $C^\ast$ is automatically tuned when using \texttt{SVM}$^{\text{light}}$ (Joachims 1999).
For dSLR, $\log_{10}(\gamma)$ is varied uniformly from $[-4,0]$, while $\lambda$ is fixed at 0.

Two oracle lines are drawn in each plot. The red line is computed by logistic regression and the purple line is computed
by SVM with $C=1000$, from an independent labeled sample of size 4000 with 1:4 class proportions (left column) or
1:1 class proportions (right column), which is transformed by the same scale as the original training set.
The red and purple oracle lines differ only slightly in the left column, but are virtually identical in the right column.
It should be noted that these oracle lines are not the optimal, Bayes decision boundary, because the log density ratio between the classes
is linear in $x_1$ but nonlinear in $x_2$ due to the different variances of $x_2$.

From these plots, we see the following comparison. First, the least penalized line (blue) from our method dSLR is much closer to the oracle lines (red and purple) than
those from the other methods, whether or not intercept adjustment is applied. This shows numerical support for Fisher consistency of our method,
given the labeled size 100 and unlabeled size 1000 reasonably large compared
with the feature dimension 2.
On the other hand, in spite of the relatively large labeled size, the lines from non-penalized logistic regression and SVM based on labeled data alone still differ noticeably from
the oracle lines. Hence this also shows that our method can exploit unlabeled data together with labeled data to actually achieve a better approximation to the oracle lines.

Second, with suitable choices of tuning parameters, some of the decision lines from existing methods can be reasonably close to the oracle lines.
In fact, such cases of good approximation can be found from the supervised methods RLR and SVM, but not from the semi-supervised methods ER and TSVM.
This indicates potentially unstable performances of ER and TSVM, particularly in the current setting where
the class proportions in unlabeled data differ from those in labeled data.
Moreover, SVM seems to perform noticeably worse in the right column, possibly due to intercept adjustment,
than in the left column, where the class proportions in test data underlying the oracle lines are identical those in labeled training data (hence a more favorable setting).


\section{EM algorithm for direct SLR} \label{sec:dSLR-EM}

We present an EM algorithm to numerically maximizer (\ref{eq:approx-obj}) for direct SLR, based on the SLR model defined by (\ref{eq:new-logit1})--(\ref{eq:new-logit3}).
We introduce the following data augmentation. Given the pooled data $\{(z_i,x_i):i=1,\dots,N\}$,
let
\begin{align}
    u_i | (z_i=j,x_i) \sim \mbox{Bernoulli} \, \left\{ \frac{\rho_j\exp(\beta_0+\beta_1^\T x_{i})} {1-\rho_j + \rho_j\exp(\beta_0+\beta_1^\T x_{i}) } \right\} .
\end{align}
Equivalently, $\{u_i:i=1,\ldots,N\}$ can be denoted as $\{u_{ji}: i=1,\ldots, n_j, j=1,2,3\}$, such that
$u_{ji}| x_{ji} \sim \mbox{Bernoulli}\, [\rho_j\exp(\beta_0+\beta_1^\T x_{i})/\{1-\rho_j + \rho_j\exp(\beta_0 +\beta_1^\T x_{i})]$ for $j=1,2,3$.
Similarly as in Section~\ref{sec:EM}, $u_{1i} = 0$ and $u_{2i}=1$ fixed, because $\rho_1 = 0$ and $\rho_2 =1$.

{\bf E-step.} The expectation of the average penalized log-likelihood from the augmented data,
given the current estimates $(\rho^{(t)},\beta^{(t)})$ is, up to an additive constant free of $(\rho,\beta)$,
\begin{align*}
    \tilde{Q}^{(t)}(\rho,\beta) =\frac{1}{N} \sumjminj &\Big[(1-\E^{(t)} u_{ji})\log(1-\rho_j) + \E^{(t)}  u_{ji} \{\log (\rho_j )+  (\beta_0+\beta_1^\T x_{ji}) \} \\
    &- \log \{ 1-\tilde{\alpha}(\rho) + \tilde{\alpha}(\rho)\exp(\beta_0+\beta_1^\T x_{ji})\}\Big] + \mbox{pen}(\rho,\beta),
\end{align*}
where $\E^{(t)}u_{ji} = \rho^{(t)}\exp({\beta_0}^{(t)}+ \beta_1^{(t)\T} x_{ji})/[1-\rho^{(t)} + \rho^{(t)}\exp({\beta_0}^{(t)}+ \beta_1^{(t)\T} x_{ji})]$.

{\bf M-step.} The next estimates $(\rho^{(t+1)}, \beta^{(t+1)})$ are obtained as a maximizer of the expected objective $\tilde{Q}^{(t)}(\rho,\beta)$.
Recall that $\kappa_Q^{(t)}(\rho,\beta,\alpha)$ defined in Section~\ref{sec:EM} is
\begin{align*}
    \kappa_Q^{(t)}  (\rho,\beta,\alpha) & = \frac{1}{N} \sumjminj \Big[ (1-\E^{(t)}  u_{ji})\log  \left\{\frac{1-\rho_j}{1-\alpha+ \alpha \exp(\beta_0+\beta_1^\T x_{ji})} \right\} \\
& \quad  + \E^{(t)} u_{ji} \log \left\{\frac{\rho_j \exp(\beta_0+ \beta_1^\T x_{ji}) }{ 1-\alpha+ \alpha \exp(\beta_0+\beta_1^\T x_{ji}) }  \right\}\Big] - \log N + \mbox{pen}(\rho,\beta) .
\end{align*}
It directly follows that $\tilde{Q}^{(t)}(\rho,\beta) = \kappa_Q^{(t)}\{\rho,\beta,\tilde\alpha(\rho)\}$ up to an additive constant. Therefore,
the expected objective $\tilde{Q}^{(t)}(\rho,\beta)$ is related to $\pQ^{(t)}(\rho,\beta)= \kappa_Q^{(t)}\{\rho,\beta,\hat\alpha(\beta)\}$ in the profile method,
in a similar manner as the average log-likelihood $\kappa\{\rho,\beta,\Tilde{\alpha}(\rho)\}$ in the SLR model is related to the average profile log-likelihood $\pl(\rho, \beta)$ in the ETM model before data augmentation.


Unfortunately, when $\rho$ is penalized with $\gamma >0$, there is no simple, closed-form expression for computing $\rho^{(t+1)}$ as in Proposition~\ref{prop:EM-formula}.
Nevertheless, we show that $\rho^{(t+1)}$ can be obtained as a solution to a simple equation, independently of $\beta^{(t+1)}$.

\begin{proposition}\label{prop:em_dslr2}
The estimate $\tilde\rho=\rho^{(t+1)}$ satisfies
\begin{align}
      \tilde\rho = \frac{\sum_{i=1}^{n_3} \E^{(t)}u_{3i}}{n_3}\frac{\psi(\tilde\rho)}{\psi(\tilde\rho) + \gamma} +  \rho^{0}\frac{\gamma}{\psi(\tilde\rho) + \gamma} , \label{eq:approx-rho}
\end{align}
where $\psi(\tilde\rho) = 1 -  n_3\tilde\rho(1- \tilde\rho) / \{ N \tilde{\alpha}(\tilde\rho) (1-\tilde{\alpha}(\tilde\rho))\} \in (0,1)$ because
$\tilde{\alpha}(\tilde\rho) (1-\tilde{\alpha}(\tilde\rho)) >  (n_3/N)\tilde\rho(1- \tilde\rho)$ for any $\tilde \rho\in(0,1)$ as shown in the proof of Proposition~\ref{prop:equiv}.
\end{proposition}

The formula (\ref{eq:approx-rho}) shows that $\tilde\rho=\rho^{(t+1)}$ implicitly remains a weighted average of the prior center $\rho^0$
and the empirical estimate  $n_3^{-1} \sum_{i=1}^{n_3} \E^{(t)}u_{3i}$, with the weight depending on $\gamma$.
If $\gamma=0$, then $\rho^{(t+1)}$ reduces to $n_3^{-1} \sum_{i=1}^{n_3} \E^{(t)}u_{3i}$ and hence the EM iterations $(\rho^{(t)},\beta^{(t)})$ coincide with those for profile SLR in Section~\ref{sec:EM}.
If $\gamma \to\infty$, then $\rho^{(t+1)}$ becomes fixed at $\rho^0$ and then $\beta^{(t)}$ converges to a maximizer of  $\kappa\{\rho^0,\beta,\Tilde{\alpha}(\rho^0)\} -\lambda \|\beta_1\|^2_2$,
the ridge estimator of $\beta$ in the
SLR model (\ref{eq:new-logit1})--(\ref{eq:new-logit3}) with $\rho=\rho^0$ fixed.
When $\rho^0$ is set to $\rho^\ell = n_2/ n$, this estimator of $\beta$ is identical to that from ridge logistic regression with labeled data only, except for an intercept shift.

In contrast, if $\rho=\rho^0$ is fixed in the EM algorithm for profile SLR, then $\beta^{(t)}$ converges to a maximizer of $\kappa\{\rho^0,\beta,\hat{\alpha}(\beta)\} -\lambda \|\beta_1\|^2_2$,
the ridge estimator of $\beta$ in the ETM model  (\ref{eq:ETM1})--(\ref{fm:exp-tilt2}).

\section{Technical details}

\subsection{Proof of Proposition~\ref{prop:equiv}}

By some abuse of notation, denote $\beta_0 + \beta_1^\T x$ as $\beta^\T x$. Let $\mathcal R$ be a fixed open set of $(\rho,\beta)$.
First, suppose that $(\tilde\rho,\tilde\beta)$ is a maximizer of  $\kappa\{ \rho, \beta, \tilde\alpha(\rho)\}$  over $\mathcal R$.
Then $\kappa \{ \rho, \beta, \hat\alpha(\beta)\} \le \kappa\{ \rho, \beta, \tilde\alpha(\rho)\} \le \kappa( \tilde\rho, \tilde\beta, \tilde\alpha(\tilde\rho)) $ for any $(\rho,\beta) \in \mathcal R$.
Denote $\tilde\alpha = \tilde\alpha(\tilde\rho)$.
To prove that $(\tilde\rho,\tilde\beta)$ is a maximizer of  $\kappa\{\rho, \beta,  \hat\alpha(\beta)\}$  over $\mathcal R$, we show that
$\tilde\alpha$ is a minimizer of  $\kappa(\tilde\rho, \tilde\beta,  \alpha)$,
which then implies that $\kappa \{ \rho, \beta, \hat\alpha(\beta)\}$ achieves a maximum value $\kappa( \tilde\rho, \tilde\beta, \tilde\alpha )$ at $(\tilde\rho,\tilde\beta)$.
Because $\kappa(\tilde\rho, \tilde\beta,  \alpha)$ is convex in $\alpha$, it suffices to show $A=0$, where
\begin{align*}
    A= \fpartial{\kappa(\tilde\rho, \tilde\beta,  \alpha)}{\alpha} \Big|_{\tilde\alpha} = \frac{1}{N} \sumjminj \frac{1-\exp( \tilde\beta^\T x_{ji})}{1- \tilde\alpha +\tilde \alpha\exp( \tilde \beta^\T x_{ji} )}.
\end{align*}
Because $(\tilde\rho,\tilde\beta)$ is a maximizer of  $\kappa\{ \rho, \beta, \tilde\alpha(\rho)\}$, the stationary condition in $(\rho,\beta_0)$ yields
\begin{align}
0 &= \fpartial{\kappa\{\rho,\beta, \tilde\alpha(\rho)\}}{\rho} \Big|_{(\tilde\rho,\tilde\beta)} \nonumber \\
&= -\frac{1}{N} \sum_{i=1}^{n_3} \frac{1-\exp( \tilde\beta^\T x_{3i})}{1-\tilde\rho + \tilde\rho\exp( \tilde\beta^\T x_{3i})}
    +\frac{n_3}{N^2} \sumjminj\frac{1-\exp(\tilde\beta^\T x_{ji})}{1- \tilde\alpha + \tilde\alpha \exp(\tilde\beta^\T x_{ji})} , \label{pf:equiv1}\\
0 &= \fpartial{\kappa\{\rho, \beta,\tilde \alpha(\rho)\}}{\beta_0}\Big|_{(\tilde\rho,\tilde\beta)} \nonumber \\
& = \frac{1}{N} \sumjminj \frac{\tilde\rho_j \exp(\tilde\beta^\T x_{ji})}{1-\tilde\rho_j + \tilde\rho_j \exp(\tilde\beta^\T x_{ji})}
 - \frac{1}{N} \sumjminj\frac{\tilde\alpha\exp(\tilde\beta^\T x_{ji})}{1-\tilde\alpha+\tilde\alpha \exp(\tilde\beta^\T x_{ji})} , \label{pf:equiv2}
\end{align}
where $\tilde\rho_1=0$, $\tilde\rho_2=1$, and $\tilde\rho_3=\tilde\rho$. Eq~(\ref{pf:equiv2}) is equivalent to
\begin{align}
0 = \frac{1}{N} \sumjminj \frac{1-\tilde\rho_j }{1-\tilde\rho_j + \tilde\rho_j \exp(\tilde\beta^\T x_{ji})}
 - \frac{1}{N} \sumjminj\frac{1-\tilde\alpha }{1-\tilde\alpha+\tilde\alpha \exp(\tilde\beta^\T x_{ji})} . \label{pf:equiv3}
\end{align}
Summing Eq~(\ref{pf:equiv1}) multiplied by $\tilde\rho (1-\tilde\rho)$ and Eq~(\ref{pf:equiv3}) gives
\begin{align*}
1- \tilde\alpha + (n_3/N) \tilde\rho(1-\tilde\rho) A - (1-\tilde \alpha) (1+\tilde \alpha A)= 0,
\end{align*}
or equivalently
\begin{align*}
\{\tilde \alpha(1-\tilde\alpha) -(n_3/N)\tilde\rho(1-\tilde\rho) \} A= 0.
\end{align*}
Because $\tilde\alpha = \sumjm (n_j/N)\tilde\rho_j$ and $t(1-t)$ is concave in $t$, Jensen's inequality implies that $\tilde\alpha(1-\tilde\alpha) \ge \sumjm (n_j/N) \tilde\rho_j (1-\tilde\rho_j) = (n_3/N) \tilde\rho(1-\tilde\rho)$. The inequality holds strictly, $\tilde \alpha(1-\tilde\alpha)  >(n_3/N)\tilde\rho(1-\tilde\rho)$, because $\tilde\rho_1 = 0 \not= \tilde\rho_2 = 1$. Hence $A=0$.

Next suppose that $(\hat\rho,\hat\beta)$ is a maximizer of  $\kappa\{ \rho, \beta, \hat\alpha(\beta)\}$  over $\mathcal R$.
Denote $\hat\alpha = \hat \alpha(\hat\beta)$. We  show that $\hat\alpha = \tilde \alpha ( \hat\rho) = \sumjm (n_j/N)\hat\rho_j$.
Because $(\hat\rho,\hat\beta, \hat\alpha)$ is a solution to the saddle-point problem (\ref{eq:saddle-point}),
the stationary condition in $\alpha$ or equivalently Eq~(\ref{eq:alpha}) gives
\begin{align}
    1 = \frac{1}{N}\sumjminj \frac{1}{1-\hat\alpha+\hat\alpha\exp(\hat\beta^\T x_{ji})}. \label{pf:equiv4}
\end{align}
The stationary condition in $(\rho, \beta_0)$ yields
\begin{align}
0 &= \fpartial{\kappa(\rho,\beta, \alpha)}{\rho} \Big|_{(\hat\rho,\hat\beta,\hat\alpha)}
= -\frac{1}{N} \sum_{i=1}^{n_3} \frac{1-\exp( \hat\beta^\T x_{3i})}{1-\hat\rho + \hat\rho\exp( \hat\beta^\T x_{3i})}, \label{pf:equiv5}\\
0 &= \fpartial{\kappa(\rho, \beta,\alpha)}{\beta_0}\Big|_{(\hat\rho,\hat\beta, \hat\alpha)} \nonumber \\
& = \frac{1}{N} \sumjminj \frac{\hat\rho_j \exp(\hat\beta^\T x_{ji})}{1-\hat\rho_j + \hat\rho_j \exp(\hat\beta^\T x_{ji})}
 - \frac{1}{N} \sumjminj\frac{\hat\alpha\exp(\hat\beta^\T x_{ji})}{1-\hat\alpha+\hat\alpha \exp(\hat\beta^\T x_{ji})} , \label{pf:equiv6}
\end{align}
where $\hat\rho_1=0$, $\hat\rho_2=1$, and $\hat\rho_3=\hat\rho$. Eq~(\ref{pf:equiv5}) implies
\begin{align}
\frac{n_j}{N} (1-\hat\rho_j) = \frac{1}{N} \sum_{i=1}^{n_j} \frac{1-\hat\rho_j}{1-\hat\rho + \hat\rho\exp( \hat\beta^\T x_{3i})}, \quad j=1,2,3. \label{pf:equiv7}
\end{align}
Eq~(\ref{pf:equiv6}) is equivalent to
\begin{align}
0 = -\frac{1}{N} \sumjminj \frac{1-\hat\rho_j }{1-\hat\rho_j + \hat\rho_j \exp(\hat\beta^\T x_{ji})}
 + \frac{1}{N} \sumjminj\frac{1-\hat\alpha }{1-\hat\alpha+\hat\alpha \exp(\hat\beta^\T x_{ji})} . \label{pf:equiv8}
\end{align}
Combining Eq~(\ref{pf:equiv4}) multiplied by $1-\hat\alpha$ and summing Eq~(\ref{pf:equiv7}) over $j=1,2,3$ and Eq~(\ref{pf:equiv8}) shows $1-\hat\alpha = \sumjm (n_j/N) (1-\hat\rho_j)$, that is,
$\hat\alpha = \sumjm (n_j/N) \hat\rho_j$.
Let $(\tilde\rho,\tilde\beta)$ be a maximizer of  $\kappa\{ \rho, \beta, \tilde\alpha(\rho)\}$  over $\mathcal R$. The preceding proof shows that
$\kappa \{ \rho, \beta, \hat\alpha(\beta)\}$ achieves the same maximum value  $\kappa( \tilde\rho, \tilde\beta, \tilde\alpha(\tilde\rho))$ as does
$\kappa\{ \rho, \beta, \tilde\alpha(\rho)\}$ over $\mathcal R$. Hence
$\kappa(\hat\rho, \hat\beta, \hat\alpha)$ as the maximum value of $\kappa \{ \rho, \beta, \hat\alpha(\beta)\}$
is also the maximum value of $\kappa\{ \rho, \beta, \tilde\alpha(\rho)\}$ over $\mathcal R$.
Because $\hat\alpha = \tilde\alpha(\hat\rho)$ and $\kappa(\hat\rho, \hat\beta, \hat\alpha) =\kappa(\hat\rho, \hat\beta, \tilde\alpha(\hat\rho))$,
this shows that $(\hat\rho,\hat\beta)$ is a maximizer of  $\kappa\{ \rho, \beta, \tilde\alpha(\rho)\}$  over $\mathcal R$.

\subsection{Proof of Proposition~\ref{prop:consistency}} \label{sec:prf-KL}

Denote
\begin{align*}
w(j,x; \rho,h) = \frac{n_j}{N} \frac{1-\rho_j + \rho_j \exp(h(x))}{ 1-\tilde \alpha(\rho)+ \tilde \alpha(\rho) \exp(h( x))}.
\end{align*}
First, we show $\kappa^*\{\rho, h, \tilde\alpha(\rho) \} \le \kappa^*\{\rho^*, h^*,  \tilde\alpha(\rho^*) \}$ for any $\rho\in(0,1)$ and $h(x)$.
By direct calculation, notice that up to an additive constant,
\begin{align*}
& \kappa^*\{\rho, h, \tilde \alpha(\rho) \}  = \int  \sumjm \frac{n_j}{N} \log\{  w(j,x; \rho,h) \} \{ 1-\rho^*_j + \rho^*_j \exp(h^*(x))\} \dd G_0(x ) \\
& = \int  \sumjm \log\{  w(j,x; \rho,h) \} w(j,x; \rho^*,h^*)  \{ 1-\tilde\alpha(\rho^*) +\tilde\alpha( \rho^*) \exp(h^*(x))\} \dd G_0(x ),
\end{align*}
where $\rho^*_1=0$, $\rho^*_2=1$, and $\rho^*_3=\rho^*$. Hence
\begin{align*}
& \kappa^*\{\rho^*, h^*,  \tilde\alpha(\rho^*) \}  - \kappa^*\{\rho, h, \tilde\alpha(\rho) \} \\
& = \int  \sumjm \log \left\{\frac{w(j,x; \rho^*,h^*)}{w(j,x; \rho,h) }\right\} w(j,x; \rho^*,h^*)  \{ 1-\tilde\alpha(\rho^*) +\tilde\alpha( \rho^*) \exp(h^*(x))\} \dd G_0(x )\\
& = \int \mbox{KL} \Big(w(\cdot,x; \rho^*,h^*) \| w(\cdot,x; \rho,h) \Big) \{ 1-\tilde\alpha(\rho^*) +\tilde\alpha( \rho^*) \exp(h^*(x))\} \dd G_0(x ) \ge 0,
\end{align*}
where $\mbox{KL}( q^* \| q ) = \sum_j q^*_j\log(q^*_j/q_j) $ is the Kullback--Leibler (KL) divergence between two probability vectors $(q^*_j)$ and $(q_j)$.

Next we show that $\min_{\alpha\in(0,1)} \kappa^*(\rho^*, h^*, \alpha) = \kappa^*\{\rho^*, h^*, \tilde \alpha(\rho^*) \} $, that is,
$\kappa^*(\rho^*, h^*, \alpha) \ge \kappa^*\{\rho^*, h^*, \tilde \alpha(\rho^*) \} $ for any $\alpha \in (0,1)$.
By direct calculation, we obtain
\begin{align*}
& \kappa^*(\rho^*, h^*, \alpha) - \kappa^*\{\rho^*, h^*, \tilde \alpha(\rho^*) \} \\
& = \int  \sumjm \frac{n_j}{N} \log \left\{ \frac{1-\tilde\alpha(\rho^*) +\tilde\alpha( \rho^*) \exp(h^*(x)) }{1-\alpha +\alpha\exp(h^*(x)) } \right\} \{ 1-\rho^*_j + \rho^*_j \exp(h^*(x))\} \dd G_0(x ) \\
& = \int  \log \left\{ \frac{1-\tilde\alpha(\rho^*) +\tilde\alpha( \rho^*) \exp(h^*(x)) }{1-\alpha +\alpha\exp(h^*(x)) } \right\} \{1-\tilde\alpha(\rho^*) +\tilde\alpha( \rho^*) \exp(h^*(x)) \} \dd G_0(x ) \ge 0,
\end{align*}
where the left hand side is the KL divergence between two probability distributions $\{1-\tilde\alpha(\rho^*) +\tilde\alpha( \rho^*) \exp(h^*(x)) \} \dd G_0(x )$ and
$\{1-\alpha +\alpha \exp(h^*(x)) \} \dd G_0(x )$.

\subsection{Proof of Proposition~\ref{prop:efficiency}}

By definition, $\hat{\beta}$ is a maximizer of $\pl(\beta) = \max_{\rho\in (0,1)} \pl(\rho,\beta)$. By abuse of notation, rewrite $(1,x^\T)^\T$ as $x$ and hence $\beta_0 + \beta_1^\T x$ as $\beta^\T x$.
Denote the log-likelihood, after rescaling, for logistic regression based on labeled data only as
\begin{align*}
    \kappa_\ell(\beta) = (N/n)\kappa \{\rho^{\ell}, \beta,\tilde{\alpha}(\rho^{\ell})\} =\frac{1}{n} \sum_{j=1}^2 \sum_{i=1}^{n_j} \log \left\{\frac{1-\rho_j + \rho_j\exp(\beta^\T x_{ji})}{1-\alpha^\ast + \tilde{\alpha}(\rho^\ell)\exp(\beta^\T x_{ji})}\right\},
\end{align*}
where $\rho_1=0$ and $\rho_2=1$ as before.
The goal is to compare the asymptotic efficiency of $\hat{\beta}$ and $\hat{\beta}^\ell$.
We use Lemmas \ref{lem:asym_prop}--\ref{lem:asym_prop3} presented later in the subsection.

For notational simplicity, assume that $n_j/N$ is fixed as a constant $0<\eta_j<1$ as $N\rightarrow \infty$.
The results can also be extended to the case where $n_j/N$ tends to a constant $0<\eta_j<1$, as in previous asymptotic analysis (Qin 1999).
Unless otherwise stated, $(\rho,\beta)$ are evaluated at the  true values $(\rho^*,\beta^*)$, and $\alpha$ is evaluated at $\alpha^*= \sum_{j=1}^m \eta_j \rho^*_j$,
where $\rho^*_1=0$, $\rho^*_2=1$, and $\rho^*_3 = \rho^*$.

By Lemma \ref{lem:asym_prop2}, it suffices to show
\begin{align}\label{eq:asympt1}
    \frac{1}{N}V^{-1} \leq \frac{1}{n}H^{-1}GH^{-1},
\end{align}
where $V$, $G$, and $H$ are from Lemma \ref{lem:asym_prop2}.
For $\partial\pl^*(\beta) / \partial\beta$ in Lemma \ref{lem:asym_prop2},
the inequality
\begin{align*}
    \Var \left\{\fpartial{\kappa_\ell(\beta)}{\beta} - HV^{-1} \fpartial{\pl^*(\beta)}{\beta} \right\} \geq 0
\end{align*}
implies
\begin{align}\label{eq:asympt2}
    n^{-1} G -  HV^{-1} E \left\{ \fpartial{\pl^*(\beta)}{\beta}\fpartial{\kappa_\ell(\beta)}{\beta^\T}\right\} -
    E \left\{ \fpartial{\kappa_\ell(\beta)}{\beta}\fpartial{\pl^*(\beta)}{\beta^\T}\right\}V^{-1} H + N^{-1} HV^{-1}H\geq 0.
\end{align}
Substituting the result of Lemma \ref{lem:asym_prop3} into Eq (\ref{eq:asympt2}) yields Eq (\ref{eq:asympt1}).

\vspace{.1in}
In the following, we present the three lemmas used above. See Sections~\ref{sec:prf-lemS1}--\ref{sec:prf-lemS3} for proofs.
Denote by $\kappa$ the function $\kappa(\rho, \beta,\alpha)$. As above,  $(\rho,\beta, \alpha)$ are evaluated at $(\rho^*,\beta^*,\alpha^*)$.

\vspace{.1in}
\begin{lemma}\label{lem:asym_prop}
(i) As $N\rightarrow\infty$,  we have
    \begin{align*}
        - \begin{pmatrix}
        \frac{\partial^2\kappa}{\partial\beta\partial \beta^\T} &  \frac{\partial^2\kappa}{\partial \beta \partial\rho}& \frac{\partial^2\kappa}{\partial \beta \partial\alpha}\\
           \frac{\partial^2\kappa}{\partial\rho\partial\beta^\T} &  \frac{\partial^2\kappa}{\partial\rho^2}& \frac{\partial^2\kappa}{\partial \rho \partial\alpha}\\
           \frac{\partial^2\kappa}{\partial\alpha\partial\beta^\T} &  \frac{\partial^2\kappa}{\partial\alpha\partial\rho}& \frac{\partial^2\kappa}{\partial\alpha^2}
        \end{pmatrix} \longrightarrow U^\dagger  = \begin{pmatrix}
        S_{11} & S_{12} & S_{13}\\
        S_{21} & s_{22} & s_{23}\\
        S_{31} & s_{32} & s_{33}
        \end{pmatrix}.
    \end{align*}
    in probability, where
    \begin{align*}
                S_{11} &= -\sum_{j=1}^3\frac{n_j}{N} \int \frac{(1-\rho_j^\ast)\rho_j^\ast\exp({\beta^\ast}^\T x)xx^\T}{1-\rho_j^\ast+\rho_j^\ast\exp({\beta^\ast}^\T x)}\dd G_0 + \int \frac{(1-\alpha^\ast)\alpha^\ast\exp({\beta^\ast}^\T x)xx^\T}{1-\alpha^\ast + \alpha^\ast\exp({\beta^\ast}^\T x)} \dd G_0,\quad\\
          S_{12} &= S_{21}^\T  = -\frac{n_3}{N} \int \frac{\exp({\beta^\ast}^\T x)x}{1-\rho^\ast+\rho^\ast\exp({\beta^\ast}^\T x)}\dd G_0, \\
           S_{13}& = S_{31}^\T  =  \int \frac{\exp({\beta^\ast}^\T x)x}{1-\alpha^\ast+\alpha^\ast\exp({\beta^\ast}^\T x)}\dd G_0,\\
           s_{22} &=  \frac{n_3}{N}\int \frac{(1-\exp({\beta^\ast}^\T x))^2}{1-\rho^\ast + \rho^\ast\exp({\beta^\ast}^\T x)}\dd G_0, \quad s_{33} = -\int \frac{(1-\exp({\beta^\ast}^\T x))^2}{1-\alpha^\ast+\alpha^\ast  \exp({\beta^\ast}^\T x)} \dd G_0,\\
           s_{23} &= s_{32} = 0.
    \end{align*}

(ii) Denote $\delta =\sumjm \eta_j {\rho_j^\ast}^2 - {\alpha^\ast}^2$. As $N\rightarrow \infty$,  $\sqrt{N}(\partial\kappa/\partial\beta^\T,\partial\kappa/\partial\rho,\partial\kappa/\partial\alpha)^\T$ converges to
multivariate normal with mean 0 and variance matrix
\begin{align*}
V^\dagger  = \begin{pmatrix}
        S_{11} - \delta S_{13}S_{31}  & S_{12} +  (n_3/N)S_{13} & -\delta S_{13}s_{33}\\
        S_{21} +   (n_3/N) S_{31}  & s_{22} &  (n_3/N)s_{33}\\
       - \delta S_{31}s_{33} &  (n_3/N)s_{33} & -s_{33} - \delta s_{33}^2
        \end{pmatrix}.
\end{align*}
\end{lemma}

\vspace{.1in}
\begin{lemma}\label{lem:asym_prop2}
(i) Under standard regularity conditions, $\sqrt{N}(\hat{\beta}-\beta^\ast)$ converges in distribution to $N(0,V^{-1})$,
with $V = \Var\{\sqrt{N}\partial \pl^*(\beta)/\partial \beta\} = S_{11} - s_{22}^{-1}S_{12}S_{21} - s_{33}^{-1}S_{13}S_{31}$, where
\begin{align*}
    \fpartial{\pl^*(\beta)}{\beta} =   \frac{\partial\kappa}{\partial\beta} - S_{12}s_{22}^{-1} \frac{\partial \kappa}{\partial \rho} -  S_{13} s_{33}^{-1} \fpartial{\kappa}{\alpha}  .
\end{align*}

(ii) Under standard regularity conditions, $\sqrt{n}(\hat{\beta}^\ell-\beta^\ast)$ converges in distribution to $N(0,H^{-1}GH^{-1})$, with
\begin{align*}
   H =- E \left\{\frac{\partial^2\kappa_{\ell}(\beta)}{\partial\beta\partial \beta^\T }\right\} = S_{11}^\ell , \quad G =  \Var \left\{\sqrt{n} \fpartial{\kappa_{\ell}(\beta)}{\beta}\right\} = S_{11}^\ell - \delta^{\ell}S^{\ell}_{12} S^{\ell^\T} _{12},
\end{align*}
where
\begin{align*}
    \delta^{\ell} &=\tilde\alpha(\rho^\ell)(1-\tilde\alpha(\rho^\ell)),\\
   S_{11}^\ell &=  \delta^{\ell}\int \frac{\exp({\beta^{\ast}}^\T)xx^\T }{1-\tilde\alpha(\rho^\ell) + \tilde\alpha(\rho^\ell)\exp({\beta^{\ast}}^\T x)}\dd G_0,\\
   S^{\ell}_{12}& =  \int \frac{\exp({\beta^\ast}^\T x)x}{1-\tilde\alpha(\rho^\ell)+\tilde\alpha(\rho^\ell)\exp({\beta^\ast}^\T x)}\dd G_0.
\end{align*}
\end{lemma}

\vspace{.1in}
\begin{lemma}\label{lem:asym_prop3}
The inner product of $\partial \pl^\ast(\beta)/\partial \beta$ and $\partial \kappa_\ell(\beta)/\partial \beta$ equals  $N^{-1}H$, i.e.,
  \begin{align*}
       E \left\{ \fpartial{\pl^*(\beta)}{\beta}\fpartial{\kappa_\ell(\beta)}{\beta^\T}\right\} = N^{-1}H.
  \end{align*}
\end{lemma}

\subsection{Proof of Lemma~\ref{lem:asym_prop}} \label{sec:prf-lemS1}

(i) We give the calculation of $S_{11}$ as an example. The remaining elements in $U^\dagger$ can be calculated in a similar way. First, direct calculation yields
\begin{align*}
 - \frac{\partial^2\kappa}{\partial\beta\partial \beta^\T} = \frac{1}{N}\sumjminj \left[-\frac{\rho_j(1-\rho_j)\exp(\beta^\T x_{ji})xx^\T}{\{1-\rho_j + \rho_j \exp(\beta^\T x_{ji})\}^2} + \frac{\alpha(1-\alpha)\exp(\beta^\T x_{ji})xx^\T}{\{1-\alpha + \alpha \exp(\beta^\T x_{ji})\}^2}\right]
\end{align*}
Because $\{x_{ji}:i=1,\dots,n_j\}$ are independent and identically drawn from \begin{align*}
    \dd P_j =(1-\rho_j^\ast )\dd G_0 + \rho_j^\ast\dd G_1 = \{1-\rho_j^\ast+\rho_j^\ast\exp(\beta^{\ast^\T} x) \}\dd G_0,
\end{align*} we obtain
\begin{align*}
    -E\left\{\frac{\partial^2\kappa}{\partial\beta\partial \beta^\T} \right\} =& \frac{1}{N}\sum_{j=1}^3 n_j E_{P_j}\left[-\frac{\rho_j^\ast(1-\rho_j^\ast)\exp(\beta^{\ast^\T} x)xx^\T}{\{1-\rho_j^\ast + \rho_j\ast \exp(\beta^{\ast^\T} x)\}^2} + \frac{\alpha^\ast(1-\alpha^\ast)\exp(\beta^{\ast^\T} x)xx^\T}{\{1-\alpha^\ast + \alpha^\ast \exp(\beta^{\ast^\T} x)\}^2}\right]\\
    =&-\sum_{j=1}^3 \frac{n_j}{N} \int\frac{\rho_j^\ast(1-\rho_j^\ast)\exp(\beta^{\ast^\T} x)xx^\T}{1-\rho_j^\ast + \rho_j^\ast \exp(\beta^{\ast^\T} x)}\dd G_0 + \int\frac{\alpha^\ast(1-\alpha^\ast)\exp(\beta^{\ast^\T} x)xx^\T}{1-\alpha^\ast + \alpha^\ast \exp(\beta^{\ast^\T} x)}\dd G_0,
\end{align*}
where the simplification in the second term on the right hand side uses
\begin{align*}
    \sum_{j=1}^{3} n_j\{1-\rho_j^\ast+\rho_j^\ast\exp(\beta^{\ast^\T} x)\} \dd G_0 = N\{1-\alpha^\ast + \alpha^\ast \exp(\beta^{\ast^\T}x)\} \dd G_0.
\end{align*}

(ii) For a vector $x\in \RR^{p+1}$, denote $x^{\otimes 2} = x x^\T$. We show the derivations of $V^\dagger_{11}$ and $V^\dagger_{13}$ as examples and the remaining elements in $V^\dagger$ can be derived similarly. First, we calculate  $V^\dagger_{11}$ as
\begin{align*}
    V^\dagger_{11} &= \Var\left\{\sqrt{N}\frac{\partial \kappa}{\partial \beta}\right\} \\
    &=\frac{1}{N} \sum_{j=1}^3 n_j\Var_{P_j}\left\{ \frac{\rho_j^\ast\exp(\beta^{\ast^\T} x)x}{1-\rho_j^\ast +\rho_j^\ast \exp(\beta^{\ast^\T}x)} -  \frac{\alpha^\ast\exp(\beta^{\ast^\T} x)x}{1-\alpha^\ast +\alpha^\ast \exp(\beta^{\ast^\T}x) } \right\}\\
    & = (\text{I}) - (\text{II}),
\end{align*}
where
\begin{align*}
    (\text{I}) &= \frac{1}{N} \sum_{j=1}^3 n_j E_{P_j}\left\{ \frac{\rho_j^\ast\exp(\beta^{\ast^\T} x)x}{1-\rho_j^\ast +\rho_j^\ast \exp(\beta^{\ast^\T}x)} -  \frac{\alpha^\ast\exp(\beta^{\ast^\T} x)x}{1-\alpha^\ast +\alpha^\ast \exp(\beta^{\ast^\T}x) } \right\}^{\otimes 2}\\
    &= \sum_{j=1}^3 \frac{n_j}{N}\int \frac{\rho_j^{\ast^2}\exp^2(\beta^{\ast^\T} x)xx^\T}{1-\rho_j^\ast +\rho_j^\ast \exp(\beta^{\ast^\T}x)}\dd G_0  - 
    \int \frac{\alpha^{\ast^2}\exp^2(\beta^{\ast^\T} x)xx^\T}{1-\alpha^\ast +\alpha^\ast \exp(\beta^{\ast^\T}x)}\dd G_0\\
        &=\sum_{j=1}^3 \frac{n_j}{N} \left\{ \int\rho_j^\ast\exp({\beta^\ast}^\T x)xx^\T\dd G_0 - \int\frac{(1-\rho_j^\ast)\rho_j^\ast\exp({\beta^\ast}^\T x)xx^\T}{1-\rho_j^\ast+\rho_j^\ast\exp({\beta^\ast}^\T x)}\dd G_0 \right\} \\
        &\qquad\qquad- \left\{ \int\alpha^\ast\exp({\beta^\ast}^\T x)xx^\T\dd G_0 - \int\frac{(1-\alpha^\ast)\alpha^\ast\exp({\beta^\ast}^\T x)xx^\T}{1-\alpha^\ast+\alpha^\ast\exp({\beta^\ast}^\T x)}\dd G_0 \right\}\\
        &= -\sum_{j=1}^3 \frac{n_j}{N}\int\frac{ (1-\rho_j^\ast)\rho_j^\ast\exp({\beta^\ast}^\T x)xx^\T}{1-\rho_j^\ast+\rho_j^\ast\exp({\beta^\ast}^\T x)}\dd G_0  + \int\frac{(1-\alpha^\ast)\alpha^\ast\exp({\beta^\ast}^\T x)xx^\T}{1-\alpha^\ast+\alpha^\ast\exp({\beta^\ast}^\T x)}\dd G_0 \\
        &= S_{11} ,
\end{align*}
and
\begin{align*}
    (\text{II})=&\frac{1}{N}\sum_{j=1}^3 n_j E^{\otimes 2}_{P_j}\left\{ \frac{\rho_j^\ast\exp(\beta^{\ast^\T} x)x}{1-\rho_j^\ast +\rho_j^\ast \exp(\beta^{\ast^\T}x)} -  \frac{\alpha^\ast\exp(\beta^{\ast^\T} x)x}{1-\alpha^\ast +\alpha^\ast \exp(\beta^{\ast^\T}x)} \right\} \\
    =&\frac{1}{N}\sum_{j=1}^3 n_j E^{\otimes 2}_{P_j} \left[\frac{(\rho_j^\ast-\alpha^\ast)\exp(\beta^{\ast^\T} x)x }{\{1-\rho_j^\ast + \rho_j^\ast \exp(\beta^{\ast^\T} x)\}\{1-\alpha^\ast + \alpha^\ast \exp(\beta^{\ast^\T} x)\}}\right]\\
    =&\frac{1}{N}\sum_{j=1}^3 n_j (\rho_j^\ast - \alpha^\ast )^2 \left\{\int  \frac{\exp(\beta^{\ast^\T} x)x}{1-\alpha^\ast + \alpha^\ast\exp(\beta^{\ast^\T} x) }\dd G_0 \right\}^{\otimes2}\\
    =& \delta S_{13}S_{31}.
\end{align*}
Hence $V^\dagger_{11} = S_{11}-\delta S_{13}S_{31}$.
Second, we calculate $V^\dagger_{13}$ as
\begin{align*}
    V^\dagger_{13} &= \Cov\left\{\sqrt{N} \frac{\partial \kappa}{\partial \beta},\sqrt{N}\frac{\partial \kappa}{\partial \alpha}\right\}\\ &=
    \frac{1}{N}\sumjm n_j \Cov_{P_j} \left\{\frac{\rho_j^\ast\exp(\beta^{\ast^\T}x)x}{1-\rho_j^\ast +\rho_j^\ast\exp(\beta^{\ast^\T}x)}  - \frac{\alpha^\ast\exp(\beta^{\ast^\T}x)x}{1-\alpha^\ast +\alpha^\ast\exp(\beta^{\ast^\T}x)}, \frac{1-\exp(\beta^{\ast^\T}x)}{1-\alpha^\ast+ \alpha^\ast \exp(\beta^{\ast^\T}x)}\right\}\\
    &= \text{(III)} - \text{(IV)},
\end{align*}
where
\begin{align*}
     \text{(III)} &= \frac{1}{N}\sumjm n_j E_{P_j} \left[  \left\{\frac{\rho_j^\ast\exp(\beta^{\ast^\T}x)x}{1-\rho_j^\ast +\rho_j^\ast\exp(\beta^{\ast^\T}x)}  - \frac{\alpha^\ast\exp(\beta^{\ast^\T}x)x}{1-\alpha^\ast +\alpha^\ast\exp(\beta^{\ast^\T}x)} \right\} \left\{ \frac{1-\exp(\beta^{\ast^\T}x)}{1-\alpha^\ast+ \alpha^\ast \exp(\beta^{\ast^\T}x)}\right\} \right]\\
     &=\frac{1}{N}\sumjm n_j E_{P_j} \left[ \frac{(\rho_j^\ast -\alpha^\ast)\exp(\beta^{{\ast}^\T}x)(1-\exp(\beta^{{\ast}^\T}x))x}{\{1-\rho_j^\ast+\rho_j^\ast \exp(\beta^{\ast^\T}x)\}\{1-\alpha^\ast+\alpha^\ast \exp(\beta^{\ast^\T}x)\}^2} \right]\\
     &=\frac{1}{N} \underbrace{\sumjm n_j(\rho_j^\ast -\alpha^\ast)}_{=0} \int \frac{\exp(\beta^{{\ast}^\T}x)(1-\exp(\beta^{{\ast}^\T}x))x}{\{1-\alpha^\ast+\alpha^\ast \exp(\beta^{\ast^\T}x)\}^2} \dd G_0\\
     &= 0 ,
\end{align*}
and
\begin{align*}
    \text{(IV)} =& \frac{1}{N}\sumjm n_j E_{P_j} \left\{ \frac{\rho_j^\ast\exp(\beta^{\ast^\T} x)x}{1-\rho_j^\ast +\rho_j^\ast \exp(\beta^{\ast^\T}x)} -  \frac{\alpha^\ast\exp(\beta^{\ast^\T} x)x}{1-\alpha^\ast +\alpha^\ast \exp(\beta^{\ast^\T}x)} \right\}E_{P_j} \left\{ \frac{1-\exp(\beta^{\ast^\T}x)}{1-\alpha^\ast +\alpha^\ast\exp(\beta^{\ast^\T}x)} \right\}\\
    =&\frac{1}{N}\sumjm n_j\left\{\int\frac{ (\rho_j^\ast -\alpha^\ast)\exp(\beta^{\ast^\T}x)x}{1-\alpha^\ast + \alpha^\ast \exp(\beta^{\ast^\T}x)}\dd G_0\right\} \left[\int \frac{(1-\exp(\beta^{\ast^\T}x))\{1-\rho_j^\ast +\rho_j^\ast\exp(\beta^{\ast^\T}x)\}}{1-\alpha^\ast +\alpha^\ast\exp(\beta^{\ast^\T}x)}\dd G_0\right]\\
    =&\frac{1}{N}\sumjm n_j (\rho_j^\ast -\alpha^\ast)S_{13}  \left[\int \frac{(1-\exp(\beta^{\ast^\T}x))\{1-\rho_j^\ast +\rho_j^\ast\exp(\beta^{\ast^\T}x)\}}{1-\alpha^\ast +\alpha^\ast\exp(\beta^{\ast^\T}x)}\dd G_0 - \underbrace{\int 1-\exp(\beta^{\ast^\T}x)\dd G_0}_{=0}\right]\\
    =&-\frac{1}{N}\sumjm n_j (\rho_j^\ast - \alpha^\ast)^2 S_{13} \int \frac{(1-\exp(\beta^{\ast^\T}x))^2}{1-\alpha^\ast + \alpha^\ast \exp(\beta^{\ast^\T}x)} \dd G_0\\
    =&\delta S_{13} s_{33}.
\end{align*}
Hence $V^\dagger_{13} = -\delta S_{13}s_{33}$.

\subsection{Proof of Lemma~\ref{lem:asym_prop2}}\label{sec:prf-lemS2}

(i) Note that $\pl(\beta) = \kappa(\rho,\beta,\alpha)$ with $\rho = \hat{\rho}(\beta)$ and $\alpha = \hat{\alpha}(\beta)$ satisfying $\partial \kappa(\rho,\beta,\alpha)/ \partial \rho = 0$ and $\partial \kappa(\rho,\beta,\alpha)/ \partial \alpha = 0$. By implicit differentiation, the gradient and Hessian of $\pl(\beta)$ are
\begin{align}
    \frac{\partial\pl}{\partial\beta} &= \frac{\partial\kappa}{\partial \beta}\bigg\rvert_{\phi=\hat{\phi}(\beta)}, \label{eq:asympt4}\\
    \frac{\partial^2\pl}{\partial\beta\partial\beta^\T} &=  \left\{\frac{\partial^2\kappa}{\partial\beta\partial\beta^\T} - \frac{\partial^2\kappa}{\partial\beta\partial\phi^\T}\left(\frac{\partial^2\kappa}{\partial\phi\partial\phi^\T}\right)^{-1}\frac{\partial^2\kappa}{\partial\phi\partial\beta^\T}\right\}\Bigg\rvert_{\phi=\hat{\phi}(\beta)},\label{eq:asympt5}
\end{align}
where $\kappa(\rho,\beta,\alpha)$ is treated as $\kappa(\beta,\phi)$ with $\phi=(\rho,\alpha)^\T$ and $\hat{\phi}(\beta) = \{\hat\rho(\beta),\hat\alpha(\beta)\}^\T$.

We use similar arguments as in the proof of Proposition 2 in Tan (2009). Write $U^{\dagger}$ as a $2\times 2$ block matrix
\begin{align*}
    U^{\dagger} = \begin{pmatrix}
        \Sigma_{11} &\Sigma_{10}\\
        \Sigma_{01}& \Sigma_{00}
     \end{pmatrix}
\end{align*}
where $\Sigma_{00}$ is the right-bottom $2\times 2$ diagonal matrix with diagonal elements $s_{22}$ and $s_{33}$.
By the asymptotic theory of M-estimators, the equation $0 =\partial \kappa/\partial \phi\rvert_{\beta=\beta^\ast}$ admits a solution $\hat{\phi}(\beta^\ast)  = \phi^\ast + o_p(N^{-1/2})$  with $\phi^*=(\rho^*,\alpha^*)^\T$.
More specifically,
\begin{align*}
    \hat{\phi}(\beta^\ast) -\phi^\ast = -\left(\frac{\partial^2\kappa}{\partial\phi\partial\phi^\T}\right)^{-1}\frac{\partial \kappa}{\partial\phi}\bigg\rvert_{\beta=\beta^\ast,\phi=\phi^\ast} + o_p(N^{-1/2}).
\end{align*}
By a Taylor expansion of ($\partial\pl/\partial\beta)(\beta^\ast)$ in Eq (\ref{eq:asympt4}), with $\hat{\phi}(\beta^\ast)$ around $\phi^\ast$, we find
\begin{align*}
    \frac{\partial \pl}{\partial \beta}\bigg\rvert_{\beta=\beta^\ast} = \left\{\fpartial{\kappa}{\beta}-\frac{\partial^2\kappa}{\partial\beta\partial\phi^\T}\left(\frac{\partial^2\kappa}{\partial\phi\partial\phi^\T}\right)^{-1}\frac{\partial \kappa}{\partial\phi}\right\}\bigg\rvert_{\beta=\beta^\ast,\phi=\phi^\ast} + o_p(N^{-1/2}).
\end{align*}
By the law of large numbers, $(\partial^2\kappa/\partial\beta\partial\phi^\T)(\beta^*,\phi^*)$ and $(\partial^2\kappa/\partial\phi\partial\phi^\T)(\beta^*,\phi^*)$ converge in probability to $\Sigma_{10}$ and $\Sigma_{00}$ respectively as $N\rightarrow\infty$.
With $\Sigma_{10}\Sigma_{00}^{-1} = (s_{22}^{-1}S_{12}, s_{33}^{-1}S_{13})$, we have $(\partial\pl/\partial\beta) (\beta^*) = (\partial\pl^*/ \partial\beta)(\beta^*) + o_p(N^{-1/2})$. Then, as $N\rightarrow \infty$, $\sqrt{N} (\partial \pl/\partial \beta) (\beta^*)$ converges to multivariate normal with mean zero and variance matrix by Lemma~\ref{lem:asym_prop}(ii),
\begin{align*}
    V &= \Var \left\{\sqrt{N}\fpartial{\pl^*}{\beta}(\beta^*) \right\}  = ( I, -\Sigma_{10}\Sigma_{00}^{-1}) V^\dagger \begin{pmatrix}
    I\\-\Sigma_{00}^{-1}\Sigma_{01}
    \end{pmatrix}  \\
    & = S_{11} - s_{22}^{-1}S_{12}S_{21} - s_{33}^{-1}S_{13}S_{31} .
\end{align*}
The simplification follows because $(I, -\Sigma_{10}\Sigma_{00}^{-1})(S_{13}^\T ,0,s_{33})^\T  =0$ and
\begin{align*}
   {V^\dagger} = { U^\dagger} - \delta \begin{pmatrix}S_{13}\\0\\s_{33}\end{pmatrix}  \begin{pmatrix}S_{13}\\0\\s_{33}\end{pmatrix}^\T  +\begin{pmatrix}S_{13}\\0\\s_{33}\end{pmatrix} \begin{pmatrix}
   0\\
   n_3/N\\
   -1
   \end{pmatrix}^\T +  \begin{pmatrix}
   0\\
   n_3/N\\
   -1
   \end{pmatrix}\begin{pmatrix}S_{13}\\0\\s_{33}\end{pmatrix} ^\T .
\end{align*}
Moreover, by Lemma \ref{lem:asym_prop}(i) and Eq (\ref{eq:asympt5}), $ -(\partial^2\pl/\partial\beta\partial\beta^\T) (\beta^*)$ converges in probability  as $N\rightarrow \infty$ to $U = \Sigma_{11}-\Sigma_{10}\Sigma_{00}^{-1}\Sigma_{01}$, which is identical to $V = S_{11} - s_{22}^{-1}S_{12}S_{21} - s_{33}^{-1}S_{13}S_{31}$. Hence $\sqrt{N}(\hat{\beta}-\beta^\ast)$ converges in distribution to $N(0,V^{-1})$.

(ii) The result follows from the sandwich variance for M-estimation and direct calculation.

\subsection{Proof of Lemma~\ref{lem:asym_prop3}}\label{sec:prf-lemS3}
By Lemma \ref{lem:asym_prop2}, we have
\begin{align}
   E \left\{ \fpartial{\pl^*(\beta)}{\beta}\fpartial{\kappa_\ell(\beta)}{\beta^\T}\right\} &= E \left\{\left(\frac{\partial\kappa}{\partial\beta} -S_{12}s_{22}^{-1}\fpartial{\kappa}{\rho}- S_{13} s_{33}^{-1} \fpartial{\kappa}{\alpha}\right)\fpartial{\kappa_\ell(\beta)}{\beta^\T}\right\}\nonumber\\
   &=E \left\{\frac{\partial\kappa}{\partial\beta}\fpartial{\kappa_\ell(\beta)}{\beta^\T}\right\} - S_{13} s_{33}^{-1}\left\{ \fpartial{\kappa}{\alpha}\fpartial{\kappa_\ell(\beta)}{\beta^\T}\right\},\label{eq:asympt3}
\end{align}
where the second equality holds because $\kappa_\ell$ is based on labeled data $\{x_{1i}\}$ and $\{x_{2i}\}$ only and hence independent of
\begin{align*}
   \fpartial{\kappa}{\rho} = \frac{1}{N} \sum_{i=1}^{n_3} \frac{-1+\exp({\beta}^\T x_{3i})}{1-\rho + \rho\exp({\beta}^\T x_{3i})}.
\end{align*}
It suffices to show that the two inner products on the right-hand side of Eq (\ref{eq:asympt3}) are
\begin{align}
    E\left\{\frac{\partial\kappa}{\partial\beta}\fpartial{\kappa_\ell(\beta)}{\beta^\T}\right\} &= N^{-1} (S^\ell_{11} - \delta^{\ell}S_{13} S^{\ell^\T}_{12}),\label{eq:asympt6}\\
    E\left\{\frac{\partial\kappa}{\partial\alpha}\fpartial{\kappa_\ell(\beta)}{\beta^\T}\right\} &=-N^{-1} ( s_{33} \delta^{\ell} S^{\ell^\T}_{12} ).\label{eq:asympt7}
\end{align}
The calculation proceeds in a similar way as in the proof of Lemma \ref{lem:asym_prop}. Because $\partial{\kappa}/\partial{\beta}$, $\partial{\kappa_\ell}/\partial{\beta}$, and $\partial{\kappa}/\partial{\alpha}$ all have means $0$ and $\{x_{ji}: i=1,\ldots,n_j\}$ are independent and identically drawn from $P_j$, we have
\begin{align*}
   N E \left\{ \fpartial{\kappa}{\beta}  \fpartial{\kappa_\ell(\beta)}{\beta^\T}\right\} &=   N\Cov\left\{ \fpartial{\kappa}{\beta} , \fpartial{\kappa_\ell(\beta)}{\beta}\right\}= \sum_{j=1}^2\frac{n_j}{n} E_{P_j}( A_j B_j ) - \sum_{j=1}^2\frac{n_j}{n} E_{P_j} ( A_j ) E_{P_j}( B_j ) ,\\
   NE\left\{\fpartial{\kappa}{\alpha} \fpartial{\kappa_\ell(\beta)}{\beta^\T}\right\} &=   N\Cov\left\{ \fpartial{\kappa}{\alpha} , \fpartial{\kappa_\ell(\beta)}{\beta}\right\} = \sum_{j=1}^{2}\frac{n_j}{n} E_{P_j}(CB_j) - \sum_{j=1}^2\frac{n_j}{n}E_{P_j} (C) E_{P_j}( B_j ),
\end{align*}
where
\begin{align*}
    A_j &= \frac{\rho_j^\ast\exp({\beta^\ast}^\T x)x}{ 1-\rho_j^\ast+ \rho_j^\ast\exp({\beta^\ast}^\T x)} - \frac{\alpha^\ast\exp({\beta^\ast}^\T x)x}{1-\alpha^\ast+\alpha^\ast\exp({\beta^\ast}^\T x)},\\
    B_j &= \frac{\rho_j^\ast\exp({\beta^\ast}^\T x)x^\T}{1-\rho_j^\ast + \rho_j^\ast\exp({\beta^\ast}^\T x)} - \frac{\tilde{\alpha}(\rho^\ell)\exp({\beta^\ast}^\T x)x^\T}{1-\tilde{\alpha}(\rho^\ell)+\tilde{\alpha}(\rho^\ell)\exp({\beta^\ast}^\T x)},\\
    C &= \frac{1-\exp({\beta^\ast}^\T x)}{1-\alpha^\ast +\alpha^\ast\exp({\beta^\ast}^\T x)}.
\end{align*}
For the first inner product, we calculate
\begin{align*}
     &\sum_{j=1}^2\frac{n_j}{n} E_{P_j}( A_j B_j ) \\
    &=\sum_{j=1}^2 \frac{n_j}{n} \int\frac{\rho_j^{\ast^2}\exp^2(\beta^{\ast^\T}x)xx^\T}{1-\rho_j^\ast + \rho_j^\ast \exp({\beta^\ast}^\T x)}\dd G_0  - \int \frac{\tilde{\alpha}(\rho^\ell)^2 \exp^2({\beta^\ast}^\T x)xx^\T}{1-\tilde{\alpha}(\rho^\ell) + \tilde{\alpha}(\rho^\ell)\exp({\beta^\ast}^\T x)}\dd G_0 \\
    &= -\sum_{j=1}^2\frac{n_j}{n} \int\frac{ (1-\rho_j^\ast)\rho_j^\ast\exp({\beta^\ast}^\T x)xx^\T}{1-\rho_j^\ast+\rho_j^\ast\exp({\beta^\ast}^\T x)}\dd G_0  + \int\frac{(1-\tilde{\alpha}(\rho^\ell))\tilde{\alpha}(\rho^\ell)\exp({\beta^\ast}^\T x)xx^\T}{1-\tilde{\alpha}(\rho^\ell)+\tilde{\alpha}(\rho^\ell)\exp({\beta^\ast}^\T x)}\dd G_0 = S_{11}^{\ell},\\
   &\sum_{j=1}^2\frac{n_j}{n}E_{P_j} (A_j) E_{P_j}( B_j )\\
    &=\sum_{j=1}^2\frac{n_j}{n}\left\{\int \frac{(\rho_j^\ast-\alpha^\ast)\exp(\beta^{\ast^\T}x)x}{1-\alpha^\ast + \alpha^\ast\exp({\beta^\ast}^\T x)}\dd G_0\right\}\left\{
    \int \frac{( \rho_j^\ast-\tilde{\alpha}(\rho^\ell))\exp({\beta^\ast}^\T x)x^\T}{1-\tilde{\alpha}(\rho^\ell) + \tilde{\alpha}(\rho^\ell)\exp({\beta^\ast}^\T x)}\dd G_0\right\}\\
   &=\sum_{j=1}^2 \frac{n_j}{n}(\rho_j^\ast-\alpha^\ast)(\rho_j^\ast-\tilde{\alpha}(\rho^\ell)) S_{13} S_{12}^{\ell^\T}=  \delta^\ell S_{13}S_{12}^{\ell^\T}.
\end{align*}
For the second inner product, we calculate
\begin{align*}
    &\sum_{j=1}^{2}\frac{n_j}{n} E_{P_j} (CB_j)\\
    &= \underbrace{\sum_{j=1}^{2}\frac{n_j}{n} (\rho_j^\ast-\tilde{\alpha}(\rho^\ell) )}_{=0} \int \frac{(1-\exp(\beta^{\ast^\T}x))\exp(\beta^{\ast^\T}x)x^\T}{\{1-\tilde{\alpha}(\rho^\ell) +\tilde{\alpha}(\rho^\ell)\exp(\beta^{\ast^\T}x)\}\{1-\alpha^\ast +\alpha^\ast\exp(\beta^{\ast^\T}x)\}}\dd G_0\\
    &=0 ,\\
    & E_{P_j}(C) = \int \frac{(1-\exp(\beta^{\ast^\T}x))\{1-\rho_j^\ast+\rho_j^\ast\exp(\beta^{\ast^\T}x)\}}{1-\alpha^\ast + \alpha^\ast \exp(\beta^{\ast^\T}x )}\dd G_0 - \underbrace{\int 1-\exp(\beta^{\ast^\T}x)\dd G_0}_{=0}\\
    &=(\alpha^\ast-\rho_j^\ast)\int\frac{(1-\exp(\beta^{\ast^\T}x))^2}{1-\alpha^\ast + \alpha^\ast\exp(\beta^{\ast^\T}x)}\dd G_0 = (\rho_j^\ast-\alpha^\ast) s_{33},\\
    &\sum_{j=1}^2 \frac{n_j}{n}E_{P_j}(C) E_{P_j}( B_j ) =  \sum_{j=1}^2\frac{n_j}{n}(\rho_j^\ast-\alpha^\ast)(\rho_j^\ast-\tilde{\alpha}(\rho^\ell))s_{33}S_{12}^{\ell^\T}=  \delta^\ell s_{33}S_{12}^{\ell^\T}.
\end{align*}
Putting the foregoing results together, we obtain Eqs (\ref{eq:asympt6}) and (\ref{eq:asympt7}).

\subsection{Proof of Proposition~\ref{prop:EM-formula}}

Similarly as in Proposition 1 in Tan (2009) or Lemma~\ref{lem:profile}, it can be shown by Jensen's inequality that
\begin{align*}
\pQ^{(t)} (\rho, \beta) = \min_{\alpha\in(0,1)}  \kappa_Q^{(t)}  (\rho,\beta,\alpha) = \kappa_Q^{(t)}  \{\rho,\beta, \hat\alpha(\beta) \},
\end{align*}
where $\hat\alpha(\beta)$ is a minimizer of  $\kappa_Q^{(t)}  (\rho,\beta,\alpha)$ over $\alpha$, satisfying Eq~(\ref{eq:alpha}).
Then $\rho^{(t+1)}$ is a maximizer of $\pQ^{(t)} (\rho, \beta) $ over $\rho$, independently of $\beta$, by direct calculation of the gradient.
Hence it suffices to show that
if and only if $\beta^{(t+1)}$ is a local (or global) maximizer of  $\kappa_Q^{(t)}  (\rho^{(t+1)},\beta,\alpha^{(t+1)})$,
then it is a local (or respectively global) maximizer of  $\kappa_Q^{(t)}  \{\rho,\beta, \hat\alpha(\beta) \}$.

By some abuse of notation, denote $\beta_0 + \beta_1^\T x$ as $\beta^\T x$. Let $\mathcal R$ be a fixed open set of $\beta$.
Suppose that $\tilde\beta$ is a maximizer of  $\kappa_Q^{(t)}  (\rho^{(t+1)},\beta,\alpha^{(t+1)})$ over $\mathcal R$.
Then $\kappa_Q^{(t)}  \{\rho^{(t+1)},\beta, \hat\alpha(\beta)\} \le \kappa_Q^{(t)}  (\rho^{(t+1)},\beta,\alpha^{(t+1)}) \le
\kappa_Q^{(t)}  (\rho^{(t+1)},\tilde \beta,\alpha^{(t+1)})$ for any $\beta \in \mathcal R$.
To prove  $\tilde\beta$ is a maximizer of  $\kappa_Q^{(t)} \{\rho^{(t+1)},\beta, \hat\alpha(\beta)\}$ over $\mathcal R$, we
show that $\alpha^{(t+1)}$ is a minimizer of $\kappa_Q^{(t)}  (\rho^{(t+1)},\tilde\beta,\alpha)$,
which then implies that $\kappa_Q^{(t)} \{\rho^{(t+1)},\beta, \hat\alpha(\beta)\}$ achieves a maximum value $\kappa_Q^{(t)}  (\rho^{(t+1)},\tilde\beta,\alpha^{(t+1)})$ at $\tilde\beta$.
Because $\tilde\beta$ is a maximizer of  $\kappa_Q^{(t)}  (\rho^{(t+1)},\beta,\alpha^{(t+1)})$, the stationary condition in $\beta_0$ yields
\begin{align*}
\sumjminj (1- \E^{(t)} u_{ji})  = \sumjminj  \frac{1-\alpha^{(t+1)}}{1-\alpha^{(t+1)} +\alpha^{(t+1)} \exp(\tilde\beta^\T x_{ji})}.
\end{align*}
Combined with the definition of $\alpha^{(t+1)}$ in (\ref{eq:EM-update}), this shows that $\alpha^{(t+1)}$ satisfies
\begin{align*}
1 = \frac{1}{N} \sumjminj  \frac{1}{1-\alpha^{(t+1)} +\alpha^{(t+1)} \exp(\tilde\beta^\T x_{ji})} ,
\end{align*}
which is the stationary condition for minimization of $\kappa_Q^{(t)}  (\rho^{(t+1)},\tilde\beta,\alpha)$, convex in $\alpha$.

Next suppose that $\hat\beta$ is a maximizer of $\kappa_Q^{(t)}  \{\rho^{(t+1)},\beta, \hat\alpha(\beta) \}$ over $\mathcal R$.
Then $\{\hat\beta, \hat\alpha(\hat\beta)\}$ is a solution to the saddle-point problem, $\max_\beta \min_\alpha \kappa_Q^{(t)}  (\rho^{(t+1)},\beta,\alpha)$.
The stationary condition in $\alpha$ gives
\begin{align*}
1 = \frac{1}{N} \sumjminj  \frac{1}{1-\hat\alpha(\hat\beta) +\hat\alpha(\hat\beta)\exp(\hat\beta^\T x_{ji})} .
\end{align*}
The stationary condition in $\beta_0$ yields
\begin{align*}
\sumjminj   (1-\E^{(t)} u_{ji})  = \sumjminj  \frac{1-\hat\alpha(\hat\beta) }{1-\hat\alpha(\hat\beta)  +\hat\alpha(\hat\beta) \exp(\hat\beta^\T x_{ji})}.
\end{align*}
These two equations together imply that $\hat\alpha(\hat\beta) = N^{-1}\sumjminj  \E^{(t)}  u_{ji} = \alpha^{(t+1)}$.
Then the stationary condition for  $\{\hat\beta, \hat\alpha(\hat\beta)\}$ to be a saddle point of $\kappa_Q^{(t)}  (\rho^{(t+1)},\beta,\alpha)$ gives
\begin{align*}
0 = \fpartial{\kappa_Q^{(t)} (\rho^{(t+1)},\beta, \alpha) }{\beta} \Big|_{(\hat\beta,\hat\alpha(\hat\beta))}
= \fpartial{\kappa_Q^{(t)} (\rho^{(t+1)},\beta, \alpha^{(t+1)}) }{\beta} \Big|_{\hat\beta}.
\end{align*}
Because $\kappa_Q^{(t)} (\rho^{(t+1)},\beta, \alpha^{(t+1)})$ is concave in $\beta$ as mentioned in Section~\ref{sec:EM},
this implies that $\hat\beta$ is a maximizer of  $\kappa_Q^{(t)} (\rho^{(t+1)},\beta, \alpha^{(t+1)})$ over $\mathcal R$.

\subsection{Proof of Proposition~\ref{prop:CEM}}

By construction, the estimate $(\beta^{c(t+1)}_0, \beta_1^{(t+1)} )$ satisfies the stationary condition
for maximization of (\ref{eq:CEM-obj2}):
\begin{align*}
0 = \sumjminj \Big[ \E^{(t)} u_{ji} - \{1+ \exp(-\beta^{c(t)}_0-\beta_1^{(t)\T} x_{ji})\}^{-1} \Big] (1, x_{ji}^\T)^\T .
\end{align*}
Let $(\beta^{c(\infty)}_0, \beta_1^{(\infty)} )$ be the limit of the sequence $(\beta^{c(t)}_0, \beta_1^{(t)})$ as $t\to\infty$. Then
$(\beta^{c(\infty)}_0, \beta_1^{(\infty)} )$ satisfies
\begin{align*}
0 &= \sumjminj \Big[ \E^{(\infty)} u_{ji} - \{1+ \exp(-\beta^{c(\infty)}_0-\beta_1^{(\infty)\T} x_{ji})\}^{-1} \Big] (1, x_{ji}^\T)^\T  \\
& = \sum_{j=1}^2 \sum_{i=1}^{n_j} \Big[ y_{ji} - \{1+ \exp(-\beta^{c(\infty)}_0-\beta_1^{(\infty)\T} x_{ji})\}^{-1} \Big] (1, x_{ji}^\T)^\T ,
\end{align*}
because $\E^{(\infty)} u_{ji} = y_{ji}$ for $j=1,2$, or $\{1+ \exp(-\beta^{c(\infty)}_0-\beta_1^{(\infty)\T} x_{ji})\}^{-1}$ if $j=3$.
This is precisely the score equation for the MLE of $(\beta^c_0,\beta_1)$ in logistic regression based on the labeled data only.

\subsection{Proof of Proposition~\ref{prop:em_dslr2}}

Rewrite $(1,x^\T)^\T$ as $x$ and $\beta_0 + \beta_1^\T x$ as $\beta^\T x$. Denote $\tilde\alpha = \tilde\alpha(\tilde\rho)$ and
\begin{align*}
    A= \frac{1}{N} \sumjminj  \frac{1-\exp(\beta^\T x_{ji})}{1-\tilde\alpha +\tilde\alpha\exp(\beta^\T x_{ji})}.
\end{align*}
Suppose $(\tilde{\rho},\tilde{\beta})$ is the maximizer of $\tilde{Q}^{(t)}(\rho,\beta)$. The stationary conditions in $(\rho,\beta_0)$ gives
\begin{align}
0 &= \fpartial{\tilde{Q}^{(t)}(\rho,\beta)}{\rho}\Big|_{(\tilde\rho,\tilde\beta)} \nonumber\\
&= \frac{1}{N}\sum_{i=1}^N  \left\{-\frac{\sum_{i=1}^{n_3}\E^{(t)}(1-u_{3i})+N\tau_1}{1-\tilde\rho} + \frac{\sum_{i=1}^{n_3}\E^{(t)}u_{3i}+N\tau_2}{\tilde\rho} + n_3 A\right\}, \label{eq:em_dslr1}\\
0 &= \fpartial{\tilde{Q}^{(t)}(\rho,\beta)}{\beta_0}\Big|_{(\tilde\rho,\tilde\beta)} \nonumber\\
&= \frac{1}{N} \left\{n_2 + \sum_{i=1}^{n_3}\E^{(t)}u_{3i} -\sumjminj \frac{\tilde{\alpha}\exp(\beta^\T x_{ji})}{1-\tilde\alpha +\tilde\alpha\exp(\beta^\T x_{ji})} \right\}, \label{eq:em_dslr2}
\end{align}
where $\tau_1 = \gamma(1-\rho^0)n_3/N$ and $\tau_2 = \gamma\rho^0n_3/N$.
Taking a difference between Eq (\ref{eq:em_dslr1}) multiplied by  $\tilde\rho(1-\tilde\rho)$ and Eq (\ref{eq:em_dslr2}) yields
\begin{align}\label{eq:em_dslr3}
    A = \frac{N\tilde\rho(\tau_1+\tau_2) - N\tau_2}{n_3\tilde\rho(1-\tilde\rho) -N\tilde\alpha(1-\tilde\alpha)}.
\end{align}
In addition, Eq (\ref{eq:em_dslr2}) is equivalent to
\begin{align}\label{eq:em_dslr4}
     A =\frac{n_3\tilde\rho - \sum_{i=1}^{n_3}\E^{(t)}u_{i3}}{N\tilde\alpha(1-\tilde\alpha)}.
\end{align}
Combining Eq (\ref{eq:em_dslr3}), Eq (\ref{eq:em_dslr4}) and the definitions of $\tau_1$ and $\tau_2$ leads to Eq~(\ref{eq:approx-rho}).

\section{Experiment details} \label{sec:experiment-details}

The 11 UCI datasets are available from \url{https://archive.ics.uci.edu/ml/datasets.php} and the 4 SSL benchmark data sets are from \url{http://olivier.chapelle.cc/ssl-book/benchmarks.html}.
Table \ref{Table:data_sets_stat} gives the statistics of the datasets.

\begin{table}[!ht]
\caption{Statistics for data sets in numerical experiments}\label{Table:data_sets_stat}
  \centering
\begin{tabular}{ccccccc}\toprule
No & Data          & \# of obs & \# of positive & \# of negative & \% of positive & feature dim \\\hline
1  & AUSTRA    & 690       & 383            & 307            & 55.51            & 14        \\
2  & BCW           & 683       & 444            & 239            & 65.01            & 9         \\
3  & GERMAN        & 1000      & 700            & 300            & 70.00            & 24        \\
4  & HEART         & 297       & 137            & 160            & 46.13            & 13        \\
5  & IONO          & 331       & 126            & 225            & 38.07            & 34        \\
6  & LIVER & 345       & 145            & 200            & 42.03            & 6         \\
7  & PIMA          & 768       & 500            & 268            & 65.10            & 8         \\
8  & SPAM          & 4601      & 2788           & 1813           & 60.60            & 57        \\
9  & VEHICLE       & 435       & 218            & 217            & 50.11            & 18        \\
10 & VOTES         & 435       & 257            & 168            & 59.08            & 16        \\
11 & WDBC          & 569       & 357            & 212            & 62.74            & 31        \\
12 & BCI           & 400       & 200            & 200            & 50.00            & 117       \\
13 & COIL          & 1500      & 750            & 750            & 50.00            & 241       \\
14 & DIGIT1        & 1500      & 766            & 734            & 51.07            & 241       \\
15 & USPS          & 1500      & 1200           & 300            & 80.00            & 241     \\\bottomrule
\end{tabular}
\end{table}

Each dataset is randomly divided into training and test data as described in Section~\ref{sec:experiment}.
For the training set including labeled and unlabeled data, each feature is standardized to have mean 0 and variance 1.
No further standardization is performed during cross validation.

The methods RLR, ER, SVM, and TSVM are implemented using the following computer packages respectively:
\begin{itemize}
\item \texttt{glmnet}, \url{https://cran.r-project.org/web/packages/glmnet/index.html},
\item \texttt{RSSL}, \url{https://cran.r-project.org/web/packages/RSSL/index.html},
\item \texttt{libsvm}, \url{https://www.csie.ntu.edu.tw/~cjlin/libsvm/}, and
\item \texttt{SVM}$^{\text{light}}$, \url{http://svmlight.joachims.org/}.
\end{itemize}
Our methods, pSLR and dSLR, are implemented using R. The codes are available from the authors upon request.

For each method, the tuning parameters are selected by 5-fold cross validation over 8 possible values as follows. The search range for each tuning parameter is determined from exploratory experiments.
\begin{itemize}
\item RLR: The objective function for RLR is $n^{-1} \ell(\beta) + \lambda \|\beta_1\|_2^2$, where $\ell(\beta)$ is the negative likelihood function for logistic regression on labeled data. Possible values for the log ridge parameter $\log_{10}(\lambda)$ are fixed uniformly from $[-5,-1]$ for UCI datasets and from $[-4,0]$ for SSL benchmark datasets.

\item ER: The objective function for ER is $N^{-1}\{\ell(\beta) + \lambda_e H(\beta)\} + \lambda \|\beta_1\|_2^2$ where $\ell(\beta)$ is the same as in RLR and $H(\beta)$ is the entropy regularizer on the unlabeled data. Possible values for $\lambda$ are fixed in the same manner as in RLR, and values for the entropy parameter $\lambda_e$ are fixed uniformly from $[0,1]$ for all datasets.

\item pSLR and dSLR: Recall that the penalty function is $\mbox{pen}(\rho,\beta) =  \lambda \|\beta_1\|_2^2  +  \gamma (1-\rho^0)  (n_3/N) \log(1-\rho) + \gamma \rho^0  (n_3/N) \log\rho.$ Possible values for the ridge parameter $\lambda$ are fixed in the same manner as in RLR and ER, and values for $\log_{10}(\gamma)$ are fixed uniformly from $[-2,2]$ for all datasets.

\item SVM: SVM solves the following optimization problem
\begin{align*}
    \min_{w,b,\xi} \ & \ \frac{1}{2}w^\T w + C\sum_{i=1}^n\xi_i\\
    \mbox{subject to}\ & \ \tilde y_i(w^\T x_i + b) \geq 1- \xi_i, \quad i =1,\dots,n ,\\
    & \ \xi_i\geq 0, \quad i =1,\dots,n ,
\end{align*}
where $\tilde y_i = 2 y_i-1 \in \{-1,1\}$ for $i=1,\ldots,n$.
Possible values for $\log_{10}(C)$ are fixed uniformly from $[-2,2]$ for all datasets.

\item TSVM: TSVM with the class balance constraint solves the following optimization problem
\begin{align*}
    \min_{w,b,\xi} \ & \ \frac{1}{2}w^\T w + C\sum_{i=1}^n\xi_i + C^\ast \sum_{i=n+1}^N \xi^\ast_i\\
    \mbox{subject to}\ & \ \tilde y_i(w^\T x_i + b) \geq 1- \xi_i,  \quad  i =1,\dots,n, \\
     & \ \tilde y_i(w^\T x_i + b) \geq 1- \xi_i^\ast, \quad  i = n+1,\dots,N,\\
    & \ \xi_i\geq 0, \quad i =1,\dots,n,  \\
    & \ \xi_i^\ast \geq 0 , \quad i = n+1,\dots,N, \\
    &\frac{1}{N-n} \sum_{i=n+1}^N \tilde y_i = \frac{1}{n}\sum_{i=1}^n \tilde y_i ,
\end{align*}
where $\tilde y_i \in \{-1,1\}$ is the predicted label for $i=n+1,\ldots,N$.
Possible values for $\log_{10}(C)$ are fixed uniformly from $[-2,2]$ for all datasets. The parameter  $C^\ast$ is automatically tuned in the implementation of  \texttt{SVM}$^{\text{light}}$.
\end{itemize}

Logistic-type methods, RLR, ER, pSLR, and dSLR, are cross validated over the binomial deviance based on the labeled data,
and SVM-type methods, SVM and TSVM, are cross validated over the accuracy. For our methods, the binomial deviance is computed on the CV test set, using the coefficient vector
$(\hat{\beta}_0+\log(n^{\text{cv}}_2/n^{\text{cv}}_1),\hat{\beta}_1^\T)^\T$, where $(n^{\text{cv}}_1,n^{\text{cv}}_2)$ are the class sizes of labeled data in the CV training set. Because the performance measures
(binomial deviance and accuracy) are based on labeled data only, the entire set of unlabeled data is used without split in training during CV for semi-supervised methods. In the case of a tie, the smaller $\lambda$ or $C$ will be selected for RLR, SVM and TSM. For ER, pSLR, and dSLR, the smaller $\lambda$ and the larger $\gamma$ or $\lambda_e$ will be selected.

\section{Additional experiment results} \label{sec:experiment-additional}

For labeled training data size 100, Table \ref{Table:labeled-100-acc-part2} presents the accuracy results where intercept adjustment is applied in the Homo Prop scheme,
but not applied in the Flip Prop scheme. Table \ref{Table:labeled-100-auc} presents the AUC results, which are not affected by whether intercept adjustment is applied.
Comparison between Tables \ref{Table:labeled-100-acc} and \ref{Table:labeled-100-acc-part2} shows that,
with intercept adjustment versus no adjustment, the accuracies of the methods, RLR, ER, pSLR, dSLR, and SVM, are decreased only slightly in the Homo Prop scheme,
but become substantially improved in the Flip Prop scheme.

For labeled training data size 25, Table \ref{Table:labeled-25-acc} presents the accuracy results similarly as in Table~\ref{Table:labeled-100-acc},
where intercept adjustment is not applied in the Homo Prop scheme,
but applied in the Flip Prop scheme. Table \ref{Table:labeled-25-acc-part2} presents the accuracy results where intercept adjustment is applied in the Homo Prop scheme,
but not applied in the Flip Prop scheme. Table \ref{Table:labeled-25-auc} presents the AUC results, which are not affected by whether intercept adjustment is applied.

\vspace{.2in}

\begin{table}[!ht]
\caption{Classification accuracy in \% (mean $\pm$ sd) on test data over 20 repeated runs, with labeled training data size 100. Subscript $_a$ indicates that intercept adjustment is applied.
Compared with Table~\ref{Table:labeled-100-acc}, intercept adjustment in applied in the Homo Prop scheme, but not in the Flip Prop scheme.}
\label{Table:labeled-100-acc-part2}
\resizebox{\columnwidth}{!}{%

\begin{tabular}{cccccc}\toprule
Homo Prop & RLR$_a$     & ER$_a$                   & pSLR$_a$             & dSLR$_a$            & SVM$_a$                           \\\midrule
AUSTRA         & 85.48 $\pm$ 1.81 & \textbf{ 85.74 $\pm$ 1.92  }        & 85.39 $\pm$ 1.89          & 85.50 $\pm$ 1.79          & 85.22$\pm$2.05         \\
BCW            & 96.31 $\pm$ 1.08          & 96.09 $\pm$ 1.07          & 96.22 $\pm$ 1.16          & 96.29 $\pm$ 1.07          & \textbf{96.62$\pm$1.00} \\
GERMAN         & 66.40 $\pm$ 3.09          & 66.73 $\pm$ 3.02          & \textbf{67.15 $\pm$ 2.84} & 66.23 $\pm$ 3.12          & 63.89$\pm$7.30         \\
HEART          & 80.26 $\pm$ 4.20          & 80.73 $\pm$ 4.39          & \textbf{80.94 $\pm$ 3.48} & 80.21 $\pm$ 4.17          & 79.69$\pm$4.37          \\
INON           & 82.08 $\pm$ 2.74          & \textbf{83.79 $\pm$ 3.00} & 81.88 $\pm$ 3.22          & 82.08 $\pm$ 2.74          & 80.67$\pm$3.73          \\
LIVER          & 62.91 $\pm$ 6.11          & 63.04 $\pm$ 6.39          & 62.91 $\pm$ 6.27          & 63.22 $\pm$ 6.32          & \textbf{64.83$\pm$3.40} \\
PIMA           & 72.97 $\pm$ 3.72          & \textbf{73.01 $\pm$ 3.51} & 72.93 $\pm$ 3.71          & 72.77 $\pm$ 3.85          & 72.40$\pm$2.55          \\
SPAM           & 88.52 $\pm$ 2.38          & \textbf{89.06 $\pm$ 2.97} & 88.60 $\pm$ 2.75          & 88.44 $\pm$ 2.34          & 86.68$\pm$4.28          \\
VEHICLE        & 93.10 $\pm$ 2.73          & 92.59 $\pm$ 2.80          & 92.45 $\pm$ 2.83          & \textbf{93.24 $\pm$ 2.82} & 92.45$\pm$3.60          \\
VOTES          & 93.59 $\pm$ 2.56          & 93.62 $\pm$ 2.37          & 93.48 $\pm$ 2.44          & 93.48 $\pm$ 2.66          & \textbf{94.00$\pm$2.57} \\
WDBC           & 95.28 $\pm$ 1.73          & 95.58 $\pm$ 1.70 & 95.58 $\pm$ 1.48          & 95.19 $\pm$ 1.78          & \textbf{95.75$\pm$2.11  }       \\
BCI            & 66.50 $\pm$ 4.06          & 65.83 $\pm$ 3.80          & 65.86 $\pm$ 4.89          & 65.86 $\pm$ 4.40          & \textbf{69.10$\pm$5.10} \\
COIL           & 78.95 $\pm$ 3.15          & 78.96 $\pm$ 3.24          & 79.07 $\pm$ 3.89          & 78.70 $\pm$ 3.42          & \textbf{80.22$\pm$2.62} \\
DIGIT1         & 89.77 $\pm$ 1.11          & 89.32 $\pm$ 2.67          & \textbf{90.07 $\pm$ 1.16} & 89.88 $\pm$ 1.10          & 89.47$\pm$1.40         \\
USPS           & 81.39 $\pm$ 4.31          & \textbf{81.91 $\pm$ 3.99} & 81.06 $\pm$ 3.95          & 81.68 $\pm$ 4.31          & 80.63$\pm$3.88         \\\midrule
              Average accuracy                & 82.23                  & \textbf{82.40}         & 82.24                  & 82.18                  & 82.11              \\
               \# within 1\% of highest& 11/15                     & \textbf{12/15}            & 11/15                     & 11/15                     & 10/15       \\
\bottomrule
\end{tabular}
}
\noindent
\resizebox{\columnwidth}{!}{%
\begin{tabular}{ccccccc}\toprule
Flip Prop              & RLR           & ER                     & pSLR                   & dSLR                   & SVM                    & TSVM         \\\midrule
AUSTRA         & 79.50 $\pm$ 3.46          & 80.87 $\pm$ 4.10          & 80.59 $\pm$ 4.66          & 82.07 $\pm$ 3.39          & \textbf{84.24 $\pm$3.38} & 71.78 $\pm$ 7.05          \\
BCW            & \textbf{97.16 $\pm$ 1.18} & 96.62 $\pm$ 1.26          & 96.87 $\pm$ 1.47          & 97.11 $\pm$ 1.23          & 96.64 $\pm$1.24        & 95.80 $\pm$ 2.80          \\
GERMAN         & 55.95 $\pm$ 5.12          & 57.99 $\pm$ 6.21          & 58.06 $\pm$ 5.77          & 59.35 $\pm$ 4.53          & \textbf{60.32 $\pm$3.64} & 57.57 $\pm$ 4.00          \\
HEART          & 72.81 $\pm$ 6.11          & 73.96 $\pm$ 4.65          & 74.06 $\pm$ 3.97          & \textbf{75.31 $\pm$ 4.45} & 70.42 $\pm$7.50          & 62.71 $\pm$ 4.34          \\
INON           & 74.17 $\pm$ 6.50          & 74.58 $\pm$ 7.23          & 72.19 $\pm$ 5.57          & 73.12 $\pm$ 5.18          & \textbf{78.75 $\pm$6.89} & 59.22 $\pm$ 8.40          \\
LIVER          & 46.30 $\pm$ 3.31          & 46.78 $\pm$ 3.63          & 47.09 $\pm$ 2.67          & 47.26 $\pm$ 3.09          & 44.22 $\pm$3.34          & \textbf{54.30 $\pm$ 2.19} \\
PIMA           & 62.40 $\pm$ 4.98          & 62.87 $\pm$ 5.61          & 63.05 $\pm$ 5.26          &63.18 $\pm$ 6.04  & \textbf{ 63.40 $\pm$4.84 }         & 61.80 $\pm$ 2.86          \\
SPAM           & 87.98 $\pm$ 2.70          & \textbf{88.04 $\pm$ 2.95} & 87.90 $\pm$ 2.49          & 87.94 $\pm$ 2.70          & 85.74 $\pm$3.90          & 87.22 $\pm$ 5.30          \\
VEHICLE        & 83.76 $\pm$ 9.50          & 80.21 $\pm$ 8.38          & 83.41 $\pm$ 8.91          & \textbf{91.07 $\pm$ 2.77} & 83.72 $\pm$7.72        & 70.62 $\pm$ 4.63          \\
VOTES          & 91.03 $\pm$ 3.00          & 91.66 $\pm$ 2.21          & \textbf{92.03 $\pm$ 2.33} & 91.90 $\pm$ 2.07          & 91.52 $\pm$3.80          & 81.03 $\pm$ 6.02          \\
WDBC           & 95.00 $\pm$ 2.11          & 95.19 $\pm$ 1.71          & 94.68 $\pm$ 2.50          & \textbf{96.20 $\pm$ 1.71} & 95.08 $\pm$1.67          & 80.23 $\pm$ 4.81          \\
BCI            & 58.53 $\pm$ 5.63          & 58.20 $\pm$ 5.67          &\textbf{ 61.58 $\pm$ 7.39}          & \textbf{61.58 $\pm$ 6.75} & 57.74 $\pm$6.92       & 60.83 $\pm$ 3.83          \\
COIL           & 61.90 $\pm$ 5.47          & 62.19 $\pm$ 6.52          & 63.13 $\pm$ 6.79          & 65.05 $\pm$ 7.79          & \textbf{71.87 $\pm$7.04} & 66.28 $\pm$ 2.90          \\
DIGIT1         & 82.11 $\pm$ 3.99          & 82.07 $\pm$ 5.11          & 85.47 $\pm$ 3.63          & \textbf{86.04 $\pm$ 3.84} & 80.92 $\pm$5.15         & 73.10 $\pm$ 2.03          \\
USPS           & 82.05 $\pm$ 3.42          & 82.29 $\pm$ 3.29          & \textbf{83.74 $\pm$ 3.21} & 83.65 $\pm$ 3.17          & 81.44 $\pm$3.08          & 64.95 $\pm$ 1.90          \\\midrule
            Average accuracy & 75.38                  & 75.57                  & 76.26                  &\textbf{77.39}               & 76.40                  & 69.83                  \\
                \# within 1\% of highest& 3/15                   & 4/15                   & 7/15                   & \textbf{11/15}                 & 7/15                   & 3/15\\  \bottomrule
\end{tabular}
} \vskip -.15in
\end{table}

\begin{table}[t]
\caption{Classification AUC in \% (mean $\pm$ sd) on test data over 20 repeated runs, with labeled training data size 100. The AUC is not affected by whether intercept adjustment is applied.}
\label{Table:labeled-100-auc}
\resizebox{\columnwidth}{!}{%
\begin{tabular}{cccccc}\toprule
Homo Prop & RLR     & ER                   & pSLR            & dSLR            & SVM                          \\\midrule
AUSTRA     & 91.22$\pm$1.70          & \textbf{91.52$\pm$1.46} & 91.23$\pm$1.62          & 91.21$\pm$1.70          & 91.09$\pm$2.03          \\
BCW        & 99.18$\pm$0.52          & 99.23$\pm$0.47          & 99.15$\pm$0.55          & 99.18$\pm$0.52          & \textbf{99.30$\pm$0.38 }  \\
GERMAN     & 72.20$\pm$3.12          & \textbf{72.49$\pm$2.99} & 72.46$\pm$2.99          & 72.20$\pm$3.13          & 69.44$\pm$10.22        \\
HEART      & 89.03$\pm$3.06          & 88.96$\pm$3.34          & 88.96$\pm$3.13          & \textbf{89.05$\pm$3.06} & 87.90$\pm$3.67        \\
INON       & 86.36$\pm$0.81          & 85.67$\pm$4.99          & \textbf{86.66$\pm$0.61} & 86.36$\pm$0.81          &83.03$\pm$5.86        \\
LIVER      & 67.98$\pm$6.07          & 67.90$\pm$6.11          & 68.24$\pm$6.07          & 68.37$\pm$6.05          & \textbf{70.91$\pm$2.58} \\
PIMA       & 79.88$\pm$3.86          & \textbf{80.03$\pm$3.84} & 79.89$\pm$3.82          & 79.83$\pm$3.83          &79.64$\pm$3.36          \\
SPAM       & 93.83$\pm$2.11          & \textbf{94.26$\pm$2.07} & 93.58$\pm$2.24          & 93.79$\pm$2.14          & 92.36$\pm$3.21         \\
VEHICLE    & \textbf{97.28$\pm$1.75} & 96.19$\pm$2.11          & 96.21$\pm$2.13          & \textbf{97.28$\pm$1.77  }        &95.86$\pm$2.97          \\
VOTES      & \textbf{98.10$\pm$1.13} & 98.03$\pm$1.08          & 98.07$\pm$1.14          & 98.05$\pm$1.17          &97.97$\pm$1.25          \\
WDBC       & 99.12$\pm$0.81          & 99.10$\pm$0.82          & 99.13$\pm$0.77          & 99.08$\pm$0.85          & \textbf{99.14$\pm$0.78} \\
BCI        & 73.17$\pm$4.33          & 72.82$\pm$3.98          & 72.45$\pm$4.69          & 72.42$\pm$4.25          & \textbf{76.11$\pm$5.18} \\
COIL       & 85.18$\pm$3.63          & 85.22$\pm$3.84          & \textbf{85.48$\pm$3.83} & 85.07$\pm$3.55          & 84.86$\pm$2.72       \\
DIGIT1     & 96.69$\pm$0.70          & 96.32$\pm$1.51          & \textbf{96.70$\pm$0.69} & 96.69$\pm$0.70          & 96.40$\pm$0.76        \\
USPS       & \textbf{86.12$\pm$2.89} & 86.28$\pm$2.82          & 86.07$\pm$2.90          & 86.10$\pm$2.87          & 83.47$\pm$4.63          \\
      \midrule

                Average AUC      & \textbf{87.69}          & 87.60                   & 87.62                   & 87.65                   & 87.17                   \\
    \# within 1\% of highest    & \textbf{13/15}          & 12/15                   & 12/15                   & \textbf{13/15}          & 9/15                    \\

\bottomrule
\end{tabular}
}
\noindent

\resizebox{\columnwidth}{!}{%
\begin{tabular}{cccccc}\toprule
Flip Prop              & RLR           & ER                     & pSLR                   & dSLR                   & SVM                        \\\midrule
AUSTRA & 91.02$\pm$1.98          & 90.91$\pm$2.12          & 91.15$\pm$2.14          & \textbf{91.24$\pm$1.96} & 90.69$\pm$2.11         \\
BCW        & 99.27$\pm$0.36          & 99.23$\pm$0.40          & 99.18$\pm$0.41          & 99.28$\pm$0.36          & \textbf{99.33$\pm$0.32} \\
GERMAN     & 74.18$\pm$2.22          & 74.42$\pm$2.33          & \textbf{75.22$\pm$2.27} & 75.17$\pm$2.09          & 74.46$\pm$2.30        \\
HEART      & 88.31$\pm$2.75          & 88.27$\pm$2.83          & \textbf{89.32$\pm$2.39} & 89.19$\pm$2.76          & 87.03$\pm$3.30        \\
INON       & \textbf{86.20$\pm$4.17} & 86.19$\pm$4.54          & 85.73$\pm$4.95          & 85.46$\pm$4.87          & 83.49$\pm$4.27       \\
LIVER      & 65.49$\pm$8.37          & 65.52$\pm$8.68          & 67.71$\pm$7.70          & \textbf{68.13$\pm$6.74} & 62.14$\pm$15.22        \\
PIMA       & 79.19$\pm$2.97          & 79.30$\pm$2.82          & \textbf{80.10$\pm$2.53} & 79.80$\pm$2.85          & 78.98$\pm$2.77         \\
SPAM       & 93.83$\pm$2.11          & \textbf{94.26$\pm$2.07} & 93.58$\pm$2.24          & 93.79$\pm$2.14          & 92.36$\pm$3.21         \\
VEHICLE    & 94.37$\pm$5.91          & 93.49$\pm$5.66          & 95.35$\pm$5.46          & \textbf{97.41$\pm$1.62} & 94.19$\pm$4.10         \\
VOTES      & 97.64$\pm$1.04          & 97.72$\pm$0.92          & \textbf{97.95$\pm$0.97} & 97.82$\pm$1.01          & 97.09$\pm$1.27         \\
WDBC       & \textbf{99.19$\pm$0.78} & 99.14$\pm$0.80          & 99.10$\pm$1.04          & \textbf{99.19$\pm$0.87 }         & 99.13$\pm$0.57        \\
BCI        & 68.90$\pm$6.06          & 68.74$\pm$6.70          & 72.94$\pm$7.56          & \textbf{73.14$\pm$7.16} & 71.10$\pm$4.87       \\
COIL       & 76.77$\pm$7.36          & 77.44$\pm$7.07          & 79.56$\pm$8.22          & \textbf{80.33$\pm$7.46} & 74.60$\pm$13.19        \\
DIGIT1     & 95.68$\pm$1.29          & 94.76$\pm$3.81          & 96.04$\pm$1.30          & \textbf{96.49$\pm$1.32} & 94.97$\pm$1.76         \\
USPS       & 89.25$\pm$2.19          & 89.39$\pm$2.11          & \textbf{90.39$\pm$1.61} & 90.00$\pm$1.96          & 88.66$\pm$2.30          \\
\midrule
            Average AUC  & 86.62                   & 86.59                   & 87.55                   & \textbf{87.76}                   & 85.88       \\
                \# within 1\% of highest  &8/15                    & 8/15                    & 14/15                   &\textbf{15/15}                   & 5/15         \\  \bottomrule
\end{tabular}
} \vskip -.15in
\end{table}

\begin{table}[t]
\caption{Classification accuracy in \% (mean $\pm$ sd) on test data over 20 repeated runs, with labeled training data size 25. Subscript $_a$ indicates that intercept adjustment is applied
similarly as in Table~\ref{Table:labeled-100-acc}.}
\label{Table:labeled-25-acc}
\resizebox{\columnwidth}{!}{%
\begin{tabular}{ccccccc}  \toprule
Homo Prop  & RLR           & ER                     & pSLR                   & dSLR                   & SVM                    & TSVM                   \\\midrule
AUSTRA    & \textbf{82.37 $\pm$ 2.47} & 81.80 $\pm$ 2.56          & 81.76 $\pm$ 3.05           & 81.35 $\pm$ 3.62          & 80.57$\pm$3.87         & 80.46 $\pm$ 4.84          \\
BCW       & 94.84 $\pm$ 2.08          & 94.76 $\pm$ 2.72          & 95.82 $\pm$ 1.65           & 94.62 $\pm$ 2.11          & 94.20$\pm$2.22          & \textbf{96.67 $\pm$ 0.81} \\
GERMAN    & 69.50 $\pm$ 2.22          & 69.76 $\pm$ 2.36          & \textbf{69.83 $\pm$ 2.10}  & 69.47 $\pm$ 2.40          & 68.03$\pm$4.02          & 63.89 $\pm$ 5.09          \\
HEART     & \textbf{79.48 $\pm$ 3.90} & 79.01 $\pm$ 4.20          & 78.80 $\pm$ 3.74           & 79.27 $\pm$ 3.97          & 78.54$\pm$4.44          & 76.72 $\pm$ 4.22          \\
INON      & \textbf{77.21 $\pm$ 5.97} & 75.33 $\pm$ 6.75          & 76.04 $\pm$ 6.77           & 76.38 $\pm$ 6.12          & 77.04$\pm$6.51          & 76.71 $\pm$ 6.98          \\
LIVER     & 57.70 $\pm$ 6.40          & 56.91 $\pm$ 5.20          & 56.78 $\pm$ 5.29           & 57.70 $\pm$ 6.25          & \textbf{60.17$\pm$6.63} & 58.04 $\pm$ 8.68          \\
PIMA      & \textbf{68.09 $\pm$ 3.61} & 67.66 $\pm$ 3.67          & 67.03 $\pm$ 3.73           & 67.83 $\pm$ 3.68          & 67.60$\pm$3.62          & 65.25 $\pm$ 4.95          \\
SPAM      & 82.76 $\pm$ 3.35          & 83.10 $\pm$ 3.67          & 83.74 $\pm$ 2.57           & 82.68 $\pm$ 3.28          & 81.28$\pm$3.77          & \textbf{85.40 $\pm$ 2.81} \\
VEHICLE   & 73.93 $\pm$ 6.71          & 73.90 $\pm$ 7.06          & 76.00 $\pm$ 7.20           & 74.31 $\pm$ 6.32          & \textbf{79.21$\pm$5.01} & 76.28 $\pm$ 8.24          \\
VOTES     & \textbf{92.10 $\pm$ 3.68} & 91.34 $\pm$ 3.63          & 91.93 $\pm$ 3.64           & 91.90 $\pm$ 3.63          & 91.62$\pm$3.36          & 92.07 $\pm$ 3.28          \\
WDBC      & 92.17 $\pm$ 3.21          & 89.94 $\pm$ 7.91          & 92.28 $\pm$ 3.83           & 91.83 $\pm$ 3.37          & 91.39$\pm$2.95          & \textbf{92.56 $\pm$ 2.51} \\
BCI       & \textbf{56.88 $\pm$ 4.98} & 54.96 $\pm$ 4.97          & 55.45 $\pm$ 5.32           & 56.32 $\pm$ 5.08          & 56.02$\pm$5.26         & 54.85 $\pm$ 4.67          \\
COIL      & 60.62 $\pm$ 5.47          & 58.63 $\pm$ 6.85          & 57.18 $\pm$ 7.21           & 61.29 $\pm$ 5.25          & 63.65$\pm$6.64          & \textbf{64.78 $\pm$ 6.83} \\
DIGIT1    & 82.04 $\pm$ 3.83          & 82.09 $\pm$ 3.83          & 83.05 $\pm$ 4.13           & 82.08 $\pm$ 3.78          & 80.39$\pm$4.82         & \textbf{84.74 $\pm$ 3.29} \\
USPS      & 81.15 $\pm$ 2.67          & 81.13 $\pm$ 2.60          & 80.89 $\pm$ 2.66           & 81.34 $\pm$ 2.29          & \textbf{81.62$\pm$2.96} & 77.57 $\pm$ 4.15          \\\midrule
Average accuracy          &76.72 &	76.02& 	76.44	& 76.56 &	\textbf{76.76}& 	76.40                \\
  \# within 1\% of highest         & \textbf{9/15}               & 6/15                   & 7/15                    & 8/15                   & 8/15                   & 7/15                   \\
\bottomrule
\end{tabular}
}
\noindent
\resizebox{\columnwidth}{!}{%
\begin{tabular}{ccccccc}\toprule
Flip Prop & RLR$_a$     & ER$_a$                   & pSLR$_a$             & dSLR$_a$            & SVM$_a$                  & TSVM          \\\midrule
AUSTRA    & \textbf{82.13 $\pm$ 3.67} & 80.22 $\pm$ 7.61          & 81.04 $\pm$ 4.68           & 81.65 $\pm$ 4.27          &77.26$\pm$ 15.46         & 68.70 $\pm$ 5.25          \\
BCW       & 94.00 $\pm$ 1.78          & 89.18 $\pm$ 18.66         & \textbf{95.11 $\pm$ 1.48}  & 94.33 $\pm$ 1.94          & 93.84$\pm$ 3.05       & 92.04 $\pm$ 5.21          \\
GERMAN    & 60.45 $\pm$ 4.89          & \textbf{60.86 $\pm$ 4.73} & 59.49 $\pm$ 5.02           & 59.58 $\pm$ 6.80          & 55.77$\pm$ 10.17        & 52.01 $\pm$ 5.18          \\
HEART     & 76.72 $\pm$ 5.84          & 76.35 $\pm$ 5.69          & 76.35 $\pm$ 5.86           & \textbf{76.72 $\pm$ 5.08} & 74.43$\pm$ 7.72      & 62.76 $\pm$ 3.76          \\
INON      & 76.21 $\pm$ 5.18          & 75.54 $\pm$ 4.46          & 74.67 $\pm$ 7.85           & \textbf{77.96 $\pm$ 5.55} &67.92$\pm$ 14.91        & 58.25 $\pm$ 2.69          \\
LIVER     & 51.83 $\pm$ 6.79          & 51.09 $\pm$ 5.94          & 52.22 $\pm$ 8.15           & 51.70 $\pm$ 10.02         & \textbf{54.04$\pm$ 8.81} & 49.00 $\pm$ 7.49          \\
PIMA      & 66.89 $\pm$ 4.01          & 66.54 $\pm$ 3.90          & 64.18 $\pm$ 4.62           & \textbf{67.79 $\pm$ 2.77} & 60.51$\pm$ 13.44         & 56.00 $\pm$ 5.18          \\
SPAM      & 81.60 $\pm$ 4.72          & \textbf{81.84 $\pm$ 4.64} & 79.02 $\pm$ 7.62           & 81.16 $\pm$ 6.90          & 80.02$\pm$ 5.32          & 62.69 $\pm$ 5.21          \\
VEHICLE   & 73.21 $\pm$ 9.39          & 73.24 $\pm$ 9.76          & \textbf{78.00 $\pm$ 10.01} & 76.14 $\pm$ 10.74         & 65.41$\pm$ 18.95         & 64.48 $\pm$ 8.52          \\
VOTES     & 88.62 $\pm$ 4.03          & 88.79 $\pm$ 3.76          & 87.93 $\pm$ 7.29           & 89.45 $\pm$ 4.70          & \textbf{90.28$\pm$ 4.16} & 77.52 $\pm$ 11.69         \\
WDBC      & 92.69 $\pm$ 2.38          & 92.89 $\pm$ 2.21          & 91.86 $\pm$ 4.80           & \textbf{93.03 $\pm$ 1.78} & 92.47$\pm$ 3.30          & 71.22 $\pm$ 5.78          \\
BCI       & 53.65 $\pm$ 5.18          & 51.39 $\pm$ 4.04          & 51.58 $\pm$ 3.54           & \textbf{54.40 $\pm$ 6.32} & 50.56$\pm$ 6.72         & 50.60 $\pm$ 4.26          \\
COIL      & 56.61 $\pm$ 5.02          & 55.70 $\pm$ 4.44          & 56.50 $\pm$ 5.49           & \textbf{57.25 $\pm$ 5.17} & 49.56$\pm$ 9.26          & 57.92 $\pm$ 5.96          \\
DIGIT1    & 77.23 $\pm$ 4.78          & 77.43 $\pm$ 4.18          & 75.13 $\pm$ 2.91           & 75.07 $\pm$ 6.05          & \textbf{80.00$\pm$ 3.79} & 70.98 $\pm$ 2.34          \\
USPS      & 73.41 $\pm$ 6.27          & 73.27 $\pm$ 6.30          & 71.04 $\pm$ 7.07           & \textbf{74.85 $\pm$ 7.85} & 71.47$\pm$ 9.44        & 62.48 $\pm$ 3.27          \\\midrule
  Average accuracy        & 73.68                  & 72.96                  & 72.94                   & \textbf{74.07}         & 70.90                  & 63.78                  \\
    \# within 1\% of highest       & 7/15                   & 4/15                   & 3/15                    & \textbf{11/15}                 & 4/15                   & 1/15
\\\bottomrule
\end{tabular}
} \vskip -.15in
\end{table}

\begin{table}[t]
\caption{Classification accuracy in \% (mean $\pm$ sd) on test data over 20 repeated runs, with labeled training data size 25. Subscript $_a$ indicates that intercept adjustment is applied.
Compared with Table~\ref{Table:labeled-25-acc}, intercept adjustment in applied in the Homo Prop scheme, but not in the Flip Prop scheme.}
\label{Table:labeled-25-acc-part2}
\resizebox{\columnwidth}{!}{%
\begin{tabular}{cccccc}\toprule
Homo Prop & RLR$_a$     & ER$_a$                   & pSLR$_a$             & dSLR$_a$            & SVM$_a$                   \\\midrule
AUSTRA         & \textbf{81.98 $\pm$ 2.37} & 81.80 $\pm$ 2.58 & 81.54 $\pm$ 2.91          & 81.22 $\pm$ 4.09          & 80.87 $\pm$ 3.38  \\
BCW            & 95.53 $\pm$ 1.81          & 95.24 $\pm$ 2.37 & \textbf{95.96 $\pm$ 1.43} & 95.36 $\pm$ 1.90          & \textbf{95.96 $\pm$ 1.39}  \\
GERMAN         & 59.94 $\pm$ 5.16          & 58.50 $\pm$ 7.01 & 57.16 $\pm$ 6.41          & \textbf{61.91 $\pm$ 4.19} & 47.36 $\pm$ 13.28 \\
HEART          & \textbf{79.64 $\pm$ 4.12} & 78.85 $\pm$ 3.99 & 79.17 $\pm$ 3.87          & 79.53 $\pm$ 4.17          & 79.06 $\pm$ 4.39 \\
INON           & \textbf{78.67 $\pm$ 5.44} & 77.08 $\pm$ 6.41 & 75.79 $\pm$ 5.73          & 77.29 $\pm$ 6.08          & 69.17 $\pm$ 17.61 \\
LIVER          & 54.91 $\pm$ 7.69          & 54.96 $\pm$ 7.99 & \textbf{55.78 $\pm$ 7.83} & 55.65 $\pm$ 8.35          & 52.91 $\pm$ 10.55  \\
PIMA           & 65.51 $\pm$ 4.41          & 64.79 $\pm$ 5.17 & 65.43 $\pm$ 5.06          & \textbf{66.04 $\pm$ 4.35} & 56.05 $\pm$ 16.86 \\
SPAM           & 82.98 $\pm$ 2.55          & 83.42 $\pm$ 3.15 & \textbf{84.18 $\pm$ 2.26} & 83.14 $\pm$ 2.74          & 82.46 $\pm$ 2.82  \\
VEHICLE        & 74.00 $\pm$ 6.49          & 73.97 $\pm$ 6.56 & 76.10 $\pm$ 6.69 & 74.86 $\pm$ 5.84          & \textbf{76.55 $\pm$ 11.76 }\\
VOTES          & \textbf{91.34 $\pm$ 3.74} & 90.72 $\pm$ 3.81 & 91.24 $\pm$ 3.53          & 91.28 $\pm$ 3.75          & 90.38 $\pm$ 4.18  \\
WDBC           & \textbf{92.78 $\pm$ 3.05} & 89.92 $\pm$ 8.05 & 92.58 $\pm$ 4.26          & 92.19 $\pm$ 3.37          &91.83 $\pm$ 2.97  \\
BCI            & \textbf{56.99 $\pm$ 4.25} & 55.49 $\pm$ 4.89 & 55.79 $\pm$ 5.39          & 56.32 $\pm$ 4.63          &50.68 $\pm$ 8.15 \\
COIL           & 60.39 $\pm$ 5.80          & 59.83 $\pm$ 6.48 & 58.36 $\pm$ 7.09          & \textbf{60.49 $\pm$ 5.98} & 58.28 $\pm$ 12.80 \\
DIGIT1         & 82.26 $\pm$ 3.95          & 82.23 $\pm$ 3.82 & \textbf{83.07 $\pm$ 4.24} & 82.08 $\pm$ 3.86          & 80.97 $\pm$ 3.71  \\
USPS           & 75.46 $\pm$ 7.80          & 74.05 $\pm$ 9.72 & 70.75 $\pm$ 8.30          & \textbf{76.22 $\pm$ 6.98} & 58.90 $\pm$ 22.80\\
      \midrule
              Average accuracy & 75.49                  & 74.72         & 74.86                  & \textbf{75.57}         & 71.43            \\
               \# within 1\% of highest & \textbf{12/15}         & 8/15          & 10/15                  & \textbf{12/15}         & 5/15           \\
\bottomrule
\end{tabular}
}
\noindent

\resizebox{\columnwidth}{!}{%
\begin{tabular}{ccccccc}\toprule
Flip Prop              & RLR           & ER                     & pSLR                   & dSLR                   & SVM                    & TSVM         \\\midrule
AUSTRA         & 69.61 $\pm$ 12.81         & 68.91 $\pm$ 14.05 & 72.50 $\pm$ 8.84           & \textbf{74.22 $\pm$ 11.25} & 72.50$\pm$15.76 & 68.70 $\pm$ 5.25          \\
BCW            & 96.00 $\pm$ 0.56          & 90.00 $\pm$ 18.88 & 95.56 $\pm$ 1.48           & \textbf{96.44 $\pm$ 0.56}  & 96.29$\pm$1.47  & 92.04 $\pm$ 5.21          \\
GERMAN         & 46.41 $\pm$ 8.80          & 42.42 $\pm$ 11.86 & 42.48 $\pm$ 12.10          & 46.13 $\pm$ 10.22          &45.54$\pm$13.13 & \textbf{52.01 $\pm$ 5.18} \\
HEART          & 57.92 $\pm$ 10.26         & 62.03 $\pm$ 9.10  & 63.54 $\pm$ 10.21          & \textbf{65.00 $\pm$ 10.69} & 65.36$\pm$ 10.97& 62.76 $\pm$ 3.76          \\
INON           & 68.50 $\pm$ 6.46          & 60.67 $\pm$ 13.56 & 61.25 $\pm$ 12.48          & \textbf{71.21 $\pm$ 7.06}  & 57.96$\pm$17.93 & 58.25 $\pm$ 2.69          \\
LIVER          & 43.13 $\pm$ 5.91          & 42.65 $\pm$ 5.65  & 43.91 $\pm$ 5.95           & 44.61 $\pm$ 8.07           & 44.26$\pm$7.27  & \textbf{49.00 $\pm$ 7.49} \\
PIMA           & 50.86 $\pm$ 12.57         & 50.72 $\pm$ 13.32 & 52.13 $\pm$ 10.49          & 53.46 $\pm$ 11.96          & 53.11$\pm$14.04 & \textbf{56.00 $\pm$ 5.18} \\
SPAM           & 72.78 $\pm$ 8.65          & 72.68 $\pm$ 9.45  & 74.06 $\pm$ 10.03          & \textbf{75.30 $\pm$ 10.74} & 70.26$\pm$11.45 & 62.69 $\pm$ 5.21          \\
VEHICLE        & 60.31 $\pm$ 10.27         & 60.45 $\pm$ 10.23 & \textbf{68.34 $\pm$ 11.82} & 67.17 $\pm$ 12.31          & 60.72$\pm$11.52  & 64.48 $\pm$ 8.52          \\
VOTES          & 85.86 $\pm$ 4.83          & 86.17 $\pm$ 4.75  & 85.31 $\pm$ 8.50           & \textbf{87.24 $\pm$ 5.85}  & 86.07$\pm$4.93  & 77.52 $\pm$ 11.69         \\
WDBC           & 91.94 $\pm$ 2.44          & \textbf{92.33 $\pm$ 2.27}  & 90.53 $\pm$ 6.26           & {92.31 $\pm$ 2.13}  & 90.75$\pm$3.23  & 71.22 $\pm$ 5.78          \\
BCI            & 52.03 $\pm$ 3.78          & 50.30 $\pm$ 3.53  & 50.34 $\pm$ 3.63           & \textbf{52.03 $\pm$ 4.24}  & 50.60$\pm$3.79  & 50.60 $\pm$ 4.26          \\
COIL           & 51.22 $\pm$ 4.34          & 50.72 $\pm$ 4.19  & 51.30 $\pm$ 4.48           & 51.33 $\pm$ 4.36           & 51.22$\pm$4.87  & \textbf{57.92 $\pm$ 5.96} \\
DIGIT1         & 65.07 $\pm$ 3.84          & 65.14 $\pm$ 3.62  & 63.45 $\pm$ 3.18           & 57.00 $\pm$ 3.54           & 66.80$\pm$5.95 & \textbf{70.98 $\pm$ 2.34} \\
USPS           & 72.24 $\pm$ 6.36          & 68.33 $\pm$ 14.46 & 66.60 $\pm$ 14.42          & \textbf{73.30 $\pm$ 8.74}  & 72.94$\pm$6.23 & 62.48 $\pm$ 3.27          \\
\midrule
            Average accuracy & 65.59                  & 64.23          & 65.42                   & \textbf{67.12}          & 65.63          & 63.78                \\
                \# within 1\% of highest & 3/15     & 1/15           & 2/15                    & \textbf{9/15}           & 3/15           & 5/15    \\  \bottomrule
\end{tabular}
}  \vskip -.15in
\end{table}

\begin{table}[t]
\caption{Classification AUC in \% (mean $\pm$ sd) on test data over 20 repeated runs, with labeled training data size 25. The AUC is not affected by whether intercept adjustment is applied.}
\label{Table:labeled-25-auc}
\resizebox{\columnwidth}{!}{%
\begin{tabular}{cccccc}\toprule
Homo Prop & RLR     & ER                   & pSLR            & dSLR            & SVM                          \\\midrule
AUSTRA  & \textbf{89.39$\pm$2.09} & 89.27$\pm$2.26          & 88.91$\pm$2.23          & 89.10$\pm$2.23           & 88.78$\pm$3.04       \\
BCW     & 99.15$\pm$0.40          & 98.97$\pm$0.82          & 99.13$\pm$0.43          & \textbf{99.19$\pm$0.36}  & 99.09$\pm$0.87           \\
GERMAN  & \textbf{61.80$\pm$5.11} & 60.80$\pm$5.14          & 59.92$\pm$5.50          & 60.76$\pm$5.52           & 48.83$\pm$15.78       \\
HEART   & {86.56$\pm$3.57} & 86.48$\pm$3.64          & 86.38$\pm$3.66          & 86.55$\pm$3.52           & \textbf{86.60$\pm$4.60  }        \\
IONO    & \textbf{78.52$\pm$6.41} & 78.03$\pm$6.92          & 75.29$\pm$9.21          & 75.54$\pm$9.25           & 69.83$\pm$19.66         \\
LIVER   & 58.47$\pm$9.90          & 57.64$\pm$10.62         & 59.06$\pm$10.52         & \textbf{59.35$\pm$10.42 }         & 55.93$\pm$14.49\\
PIMA    & 69.94$\pm$8.03          & 69.87$\pm$8.38          & 69.33$\pm$8.48          & \textbf{70.03$\pm$8.09}  & 58.31$\pm$22.00          \\
SPAM    & 90.96$\pm$2.14          & \textbf{91.01$\pm$2.77} & 90.31$\pm$2.73          & 90.89$\pm$2.18           & 89.99$\pm$2.84           \\
VEHICLE & 79.24$\pm$6.76          & 79.50$\pm$6.94          & \textbf{81.45$\pm$6.18} & 80.71$\pm$6.46           & 80.60$\pm$17.10          \\
VOTES   & 96.88$\pm$2.07          & \textbf{97.09$\pm$1.51} & 96.96$\pm$1.89          & 96.91$\pm$2.03           & 97.03$\pm$1.79           \\
WDBC    & \textbf{97.98$\pm$1.38} & 96.87$\pm$4.01          & 97.87$\pm$1.46          & 97.94$\pm$1.43           & 97.49$\pm$1.53          \\
BCI     & 60.30$\pm$5.44          & 57.76$\pm$5.74          & 58.30$\pm$6.41          & \textbf{60.75$\pm$5.50}  & 50.98$\pm$10.58          \\
COIL    & 64.29$\pm$5.79          & 63.11$\pm$5.74          & 62.40$\pm$5.88          & \textbf{64.61$\pm$5.45}  & 60.35$\pm$14.91          \\
DIGIT1  & 90.85$\pm$3.84          & 90.86$\pm$3.87          & \textbf{91.26$\pm$4.09} & 90.85$\pm$3.83           & 90.16$\pm$3.85          \\
USPS    & \textbf{74.09$\pm$6.85} & 73.46$\pm$6.86          & 73.93$\pm$6.44          & 74.05$\pm$6.82           & 60.99$\pm$23.77        \\
      \midrule

                Average AUC      & \textbf{79.89}          & 79.38                   & 79.37                   & 79.82                    & 75.66                    \\
    \# within 1\% of highest     & \textbf{13/15}             & 9/15                       & 10/15                      & 12/15                       & 6/15                       \\
\bottomrule
\end{tabular}
}
\noindent

\resizebox{\columnwidth}{!}{%
\begin{tabular}{cccccc}\toprule
Flip Prop              & RLR           & ER                     & pSLR                   & dSLR                   & SVM                        \\\midrule
AUSTRA  & 88.78$\pm$2.28          & 88.31$\pm$2.81          & 88.54$\pm$2.81          & \textbf{88.81$\pm$2.70}  & 83.97$\pm$17.95         \\
BCW     & 99.23$\pm$0.43          & 99.16$\pm$0.45          & 98.78$\pm$1.17          & 99.21$\pm$0.41           & \textbf{99.37$\pm$0.36}  \\
GERMAN  & \textbf{66.70$\pm$4.43} & 66.50$\pm$4.41          & 66.43$\pm$4.90          & 66.42$\pm$4.44           & 58.80$\pm$14.71         \\
HEART   & 85.28$\pm$6.01          & 85.20$\pm$6.20          & 85.15$\pm$5.28          & \textbf{85.40$\pm$5.49}  & 82.45$\pm$7.68        \\
IONO    & 77.33$\pm$5.53          & 77.08$\pm$5.39          & 76.91$\pm$6.19          & \textbf{77.74$\pm$5.76}  & 67.38$\pm$21.01         \\
LIVER   & 53.70$\pm$9.55          & 52.94$\pm$8.56          & 53.78$\pm$11.19         & \textbf{54.34$\pm$12.23} & 53.70$\pm$11.89          \\
PIMA    & 70.45$\pm$6.55          & 69.72$\pm$6.97          & 70.54$\pm$5.94          & \textbf{71.64$\pm$5.21}  & 62.14$\pm$18.14          \\
SPAM    &\textbf{ 90.41$\pm$2.58}          & 90.30$\pm$2.37          & 88.22$\pm$5.26          & {90.40$\pm$3.19}  & 88.00$\pm$3.69          \\
VEHICLE & 79.18$\pm$8.79          & 78.82$\pm$9.53          & \textbf{82.57$\pm$9.94} & 81.04$\pm$10.07          & 68.72$\pm$25.54         \\
VOTES   & 96.20$\pm$1.71          & 96.39$\pm$1.54          & 95.56$\pm$2.77          & \textbf{96.36$\pm$1.98}  & 96.12$\pm$1.78          \\
WDBC    & 98.18$\pm$1.35          & \textbf{98.22$\pm$1.18} & 98.12$\pm$1.71          & 98.16$\pm$1.26           & 97.85$\pm$1.67           \\
BCI     & \textbf{56.50$\pm$5.88} & 54.00$\pm$4.47          & 53.53$\pm$4.83          & 55.74$\pm$7.26           & 49.58$\pm$8.75           \\
COIL    & \textbf{59.49$\pm$6.30 }         & 57.67$\pm$6.05          & 57.73$\pm$6.41          & 59.34$\pm$6.19  & 48.33$\pm$10.57         \\
DIGIT1  & 89.97$\pm$2.97          & \textbf{90.06$\pm$2.72} & 89.46$\pm$3.02          & 88.38$\pm$3.43           & 88.68$\pm$3.59           \\
USPS    & 79.73$\pm$3.68          & 79.50$\pm$3.70          & 80.16$\pm$3.50          & \textbf{80.46$\pm$3.39}  & 76.85$\pm$13.65                 \\
\midrule
            Average AUC  & 79.41                   & 78.92                   & 79.03                   & \textbf{79.56}           & 74.80     \\
                \# within 1\% of highest & \textbf{13/15}             & 10/15                      & 11/15                      & \textbf{13/15}              & 4/15      \\  \bottomrule
\end{tabular}
}  \vskip -.15in
\end{table}

\end{document}